\documentclass[conference]{IEEEtran}
\IEEEoverridecommandlockouts

\usepackage{cite}
\usepackage{amsmath,amssymb,amsfonts}
\usepackage{algorithmic}
\usepackage{graphicx}
\usepackage{textcomp}
\usepackage{xcolor}
\usepackage{pdfpages}
\usepackage{wrapfig}
\usepackage{url}
\def\BibTeX{{\rm B\kern-.05em{\sc i\kern-.025em b}\kern-.08em
    T\kern-.1667em\lower.7ex\hbox{E}\kern-.125emX}}

\usepackage[normalem]{ulem}
\useunder{\uline}{\ul}{}
\usepackage{graphicx}
\usepackage{multirow}
\usepackage{authblk}
\begin{document}

\title{Examining Imbalance Effects on Performance and Demographic Fairness of Clinical Language Models}

\author[1]{Precious Jones}
\author[1]{Weisi Liu}
\author[2]{I-Chan Huang}
\author[1]{Xiaolei Huang}
\affil[1]{Department of Computer Science, University of Memphis}
\affil[2]{Department of Epidemiology and Cancer Control, St Jude Children's Research Hospital}

\maketitle

\begin{abstract}
Data imbalance is a fundamental challenge in applying language models to biomedical applications, particularly in ICD code prediction tasks where label and demographic distributions are uneven. While state-of-the-art language models have been increasingly adopted in biomedical tasks, few studies have systematically examined how data imbalance affects model performance and fairness across demographic groups. This study fills the gap by statistically probing the relationship between data imbalance and model performance in ICD code prediction. 
We analyze imbalances in a standard benchmark data across gender, age, ethnicity, and social determinants of health by state-of-the-art biomedical language models. 
By deploying diverse performance metrics and statistical analyses, we explore the influence of data imbalance on performance variations and demographic fairness. 
Our study shows that data imbalance significantly impacts model performance and fairness, but feature similarity to the majority class may be a more critical factor. We believe this study provides valuable insights for developing more equitable and robust language models in healthcare applications\footnote{We released our code: \url{https://github.com/trust-nlp/ImbalanceAssessment}}.
\end{abstract}


\section{Introduction}

Data imbalance is a common yet unresolved challenge in building classifiers to support health decision making when the data has uneven distributions.
The uneven distributions can exist in various forms in health data, such as tokens, data sources, document class, and patient populations.
For example, there are more non-medical tokens than medical tokens in radiology reports~\cite{wu2023token}, and medical codes can have a skewed distribution~\cite{johnson2023mimic}.
Phenotype inference is an important classification task in healthcare, such as ICD code prediction, suffering from code class imbalance to build accurate classifiers~\cite{cloutier2023fine}.
The International Classification of Diseases (ICD) is a comprehensive coding system (e.g., over 71K codes in ICD-10-PCS) to categorize diseases, symptoms, and health-related conditions. 
While increasing studies~\cite{wornow2023shaky} have deployed clinical language models to achieve state-of-the-art performance across diverse downstream tasks, very few studies have systematically examined the effects of data imbalance on those clinical language models.

Health data collected from patients thus centers at patients and contains rich patient attributes, such as demography and social determinants of health (SDoH), which naturally root with imbalance patterns.
For example, our data analysis in Section~\ref{sec:data} on a standard data benchmark shows varying imbalance patterns on demographic groups and their subgroups, such as Hispanic/Latino female patients.
Unfortunately, existing studies~\cite{wang2020imbalanced, lu2022comparative, angeli2022class, cloutier2023fine, henning2023survey} primarily focus on class imbalance and usually leave the other imbalance factors (e.g., demography) overlooked.
The demographic and SDoH factors have been demonstrated their strong associations with patient outcomes and health disparities~\cite{kino2021scoping, roosli2022peeking, yang2023evaluating}.
However, how the demographic and SDoH imbalance patterns (e.g., racial/ethnicity) may impact clinical language models remain an unsolved question.

In this study, we will fill the gap by analyzing a fundamental task utilizing clinical language models across diverse imbalance factors.
The following three questions drive our analysis throughout the experiments:
\begin{itemize}
    \item To what extent does imbalance exist in benchmark datasets for ICD code prediction, both in terms of demographic variables and label distribution?
    \item If present, how does this imbalance influence performance disparities across demographic groups?
    \item What patterns emerge in the relationship between data imbalance and model performance and fairness across different groups of patient demography and SDoH?
\end{itemize}
To answer these questions, we include a standard benchmark dataset for ICD code prediction~\cite{johnson2023mimic-note}, examining imbalances across gender, age, ethnicity or race, and insurance status (SDoH). We include and evaluate three state-of-the-art Biomedical Language Models: ClinicalBERT~\cite{alsentzer2019publicly}, GatorTron~\cite{yang2022gatortron}, and Clinical Longformer~\cite{li2022clinical}. Our analysis explores whether performance and demographic fairness issues are consistent across all language models and if they follow similar patterns.
Furthermore, we conduct statistical analyses to investigate the correlation between data imbalance and model performance. 
To the best of our knowledge, this is the \textbf{\underline{first systematic study}} of data imbalance in medical code prediction and its impact on performance and fairness across demographic and SDoH groups. 
We expect that our findings can provide valuable insights for the health informatics community utilizing clinical language models for diverse applications, contributing to the development of more equitable and robust models in healthcare.

\begin{table*}[htp]
\centering
\caption{Insurance and Age proportions by the intersection of Race/Ethnicity and Gender}
\label{tab: subgroup_analysis}
\begin{tabular}{l|ccc|cccccc|c}
                & \multicolumn{3}{c|}{Insurance Proportion (\%)} & \multicolumn{6}{c|}{Age Proportion (\%)}                   & Average     \\
Subgroups & Medicaid & Medicare & Other & Median & 18-29 & 30-49 & 50-69 & 70-89 & 90+ &  Label Count \\ \hline\hline
Hispanic/Latino-F        & 25.67    & 24.82    & 49.52   & 52 & 13.54 & 29.64 & 35.48 & 20.29 & 1.05 & 5.08 \\
Hispanic/Latino-M        & 18.96    & 29.47    & 51.56   & 51 & 8.67  & 35.33 & 39.67 & 15.67 & 0.66 & 5.39 \\
Asian-F                  & 15.38    & 21.78    & 62.84   & 57 & 11.22 & 23.84 & 34.62 & 27.38 & 2.93 & 3.92 \\
Asian-M                  & 17.48    & 22.80    & 59.72   & 62 & 7.78  & 18.05 & 39.14 & 32.93 & 2.09 & 4.99 \\
Black/African American-F & 13.66    & 36.09    & 50.25   & 56 & 9.54  & 25.46 & 37.73 & 24.71 & 2.56 & 5.82 \\
Black/African American-M & 12.45    & 36.26    & 51.29   & 57 & 6.68  & 24.88 & 45.75 & 21.48 & 1.21 & 6.10 \\
White-F                  & 4.87     & 46.47    & 48.66   & 63 & 6.26  & 17.85 & 36.00 & 33.97 & 5.92 & 5.62 \\
White-M                  & 5.80     & 43.71    & 50.50   & 63 & 4.55  & 15.74 & 44.14 & 32.29 & 3.28 & 6.16 \\
Other/Not reported-F     & 9.53     & 34.01    & 56.46   & 60 & 9.29  & 23.16 & 34.15 & 29.20 & 4.20 & 5.41 \\
Other/Not reported-M     & 9.85     & 31.43    & 58.73   & 59 & 8.64  & 19.75 & 41.79 & 27.12 & 2.71 & 5.66
\end{tabular}
\end{table*}

\section{Related Work}

\subsection{Automatic ICD Coding}
Automatic ICD coding is a challenging task in healthcare informatics.  
This process has traditionally been costly, time-consuming. 
Recently, the application of large language models (LLMs) has led to significant advancements in automated ICD coding systems, demonstrating enhanced ability to capture complex patterns in clinical narratives \cite{biseda2020predictionicdcodesclinical}.
Despite these advancements, a common challenge faced by these models is the ICD distributional imbalance, which results in varying performance across different codes. Studies have observed that model performance tends to decline for ICD codes that are less frequent or rare \cite{ji2021does}. 
\cite{tsai2019leveraging} leveraged hierarchical category knowledge by incorporating high-level codes and additional loss terms to help models learn general concepts for low-level codes. 
\cite{pmlr-v106-xu19a} assigned higher weights to minor class samples, and \cite{jin2023learning} reweighted imbalances between positive and negative samples by positive-unlabeled learning. 
However, \textit{there remains a lack of comprehensive study on examining how imbalance effects impact model performance} across different demographic groups. 
This study aims to address this gap by systematically analyzing data imbalance effects on biomedical models and thus offering insights for future strategies of mitigating label and demographic imbalance effects.

\subsection{Model Robustness}
Model robustness, the ability to maintain consistent performance across varying conditions, remains a critical yet incompletely resolved challenge in the medical field, which contains heterogeneous data and diverse patient population. 
The emergence of large language models has been successfully adapted to various biomedical tasks~\cite{nerella2024transformers}, such as diagnosis~\cite{kuroiwa2023potential, liu2024timemattersexaminetemporal,lyu2022multimodal}, question answering~\cite{li2024dalk}, and biological reasoning~\cite{hsu2024thought}.
However, these models still exhibit vulnerabilities to various perturbations and biases.
\textit{The inherent imbalance in biomedical data} poses a unique challenge to the robustness of language models. 
For instance, \cite{wu2023token} demonstrated that token imbalance can lead to underfitting on medical tokens and reduce the quality of radiology reports, and \cite{shi2022improving} explored imbalance patterns in FDA drug datasets, revealing overfitting issues of clinical language models on majority labels.
Yet, a comprehensive understanding of how data imbalance affects model performance and fairness across different patient groups remains elusive.
Our study aims to address this gap by examining imbalance patterns in benchmark datasets, analyzing model performance across diverse demographic and SDoH groups, and conducting statistical analyses to elucidate the relationship between performance metrics and data characteristics.

\section{Data}
\label{sec:data}
We collected the clinical notes and ICD from the Medical Information Mart for Intensive Care IV Notes v2.2
~\cite{johnson2023mimic-note,goldberger2000physiobank} (MIMIC-IV), a publicly available collection of de-identified clinical notes, including discharge summary and radiology reports.
Our study experimented with the phenotype inference task that predicts International Classification of Diseases 10 (ICD-10) codes by the discharge summary.

The corpus contains 331,794 de-identified discharge summaries from 145,915 patients admitted to the hospital and emergency department at the Beth Israel Deaconess Medical Center in Boston, MA, USA.
We sourced patient demographic attributes (e.g., gender and ethnicity/race) and insurance status from MIMIC-IV~\cite{johnson2023mimic} and preprocessed the discharge summary by tokenization, code version convert, and integration. 
Appendix\ref{sub:appdix:data} include more preprocessing details for reproduction purposes.
While imbalance naturally exists, there is \textbf{no prior study} that systematically examined imbalance patterns of the MIMIC-IV data.
Close studies\cite{angeli2022class, lu2022comparative, wang2020imbalanced} mainly focus on label imbalance, while ignoring the other imbalance patterns and the fundamental cause of imbalance, diverse patient demography and their subgroups.
Thus, to better understand imbalance effects, we probe into the data imbalance patterns from three perspectives, ICD-10 codes, demographic and SDoH groups, and their subgroups (e.g., Hispanic/Latino patients with split by insurance types).

\begin{figure*}[htp]
\centering
\includegraphics[width=1\textwidth]{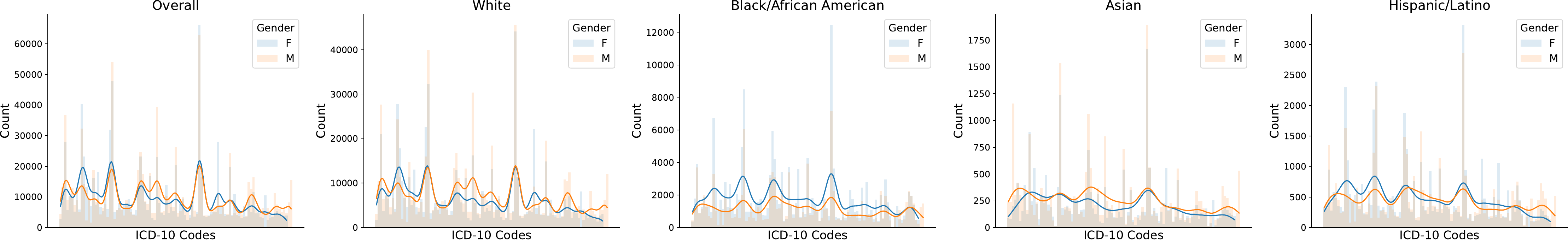}
\caption{Overview of label distribution (ICD-10 codes) by ethnicity group. The codes are arranged in descending order of frequency based on the overall data.}
\label{fig: icd_distribution}
\end{figure*} 

\subsection{Varying Demographic Imbalance}

Our dataset exhibits significant skew patterns across various demographic attributes, notably in race, age, and label distributions, shown in Figure~\ref{fig: icd_distribution} and Table~\ref{tab: subgroup_analysis}.
While gender appears relatively balanced overall, imbalances emerge within specific race / ethnicity groups. 
The detailed tables and analysis of single demographic characteristic imbalances are provided in the Appendix\ref{sub:appdix:imbalance}. 
In this section, we focus on \emph{cross-demographic characteristics} to uncover deeper insights of gender and racial intersections, insurance coverage, age, and ICD-10 codes.

\paragraph{Imbalance patterns can vary significantly across intersectional groups}
Table~\ref{tab: subgroup_analysis} shows that the intersectional groups of gender and race unveils notable disproportions in insurance coverage, leading to the data and label imbalance.
For example, the majority group, White patients, has the highest proportion of Medicare coverage among all ethnicity/racial groups, with 46.47\% of females and 43.71\% of males. 
In contrast, minority groups exhibit different imbalance patterns. 
Hispanic or Latino females have higher Medicaid coverage (25.67\%) than males (18.96\%), yet despite having a higher proportion of elderly individuals aged 70 and above (20.29\% vs. 15.67\% for males), they have lower Medicare coverage (24.82\% vs. 29.47\%).
This discrepancy might suggests that elderly Hispanic or Latino females face challenges in accessing Medicare benefits, possibly due to socioeconomic factors or lack of support.
Furthermore, Black or African American females have slightly higher Medicaid coverage (13.66\%) than males (12.45\%), which contrasts with the general trend in other group where males typically have higher Medicaid enrollment. 
The variations in insurance coverage across gender and race highlight potential causes of data imbalance distributions across demographic groups, which could indicate potential health inequality and impact the fairness of predictive models trained on this data.

\paragraph{Minority groups have younger population structures}
The age distribution across gender and race reveals notable differences (Table~\ref{tab: subgroup_analysis}). 
Hispanic or Latino patients are generally younger, with median ages of 52 years for females and 51 years for males. 
A significant proportion of Hispanic or Latino females (43.18\%) are under the age of 50, compared to 44.00\% of males. 
This contrasts sharply with the White patients, where the median age is 63 years for both genders, and over 70\% are aged 50 and above.
Among Asian individuals, females have a lower median age (57 years) compared to males (62 years), a pattern not as pronounced in other racial groups. 
These imbalances show that minority groups are underrepresented in older age brackets within the dataset. 
The lack of sufficient representation of older individuals in minority populations could result in models that are less accurate for these groups, impacting both fairness and overall model effectiveness, which is examined in this study.

\paragraph{Imbalance patterns of health diagnosis records vary significantly across gender and races}
Table~\ref{tab: subgroup_analysis} shows that the average label count—a proxy for the number of diagnosis records health conditions—reveals that males have higher counts across all racial groups. 
For instance, White males have an average count of 6.16 ICD-10 codes, compared to 5.62 for females. 
Similarly, Black or African American males have an average of 6.10 ICD-10 codes, slightly higher than the 5.82 for females.
The varying patterns may suggest the demographic factors are critical contributions to the data imbalance.
Thus, understanding these nuances is essential, as they could influence the performance of predictive models. 
If models are trained on data reflecting variations and disparities, they may perpetuate biases and lead to unequal outcomes.

\paragraph{Uneven disease prevalence: ICD-10 code variations across demography}
While existing studies only focus on class imbalance, our analysis uncovers such imbalance patterns can significantly vary across gender and race combinations (e.g., Hispanic/Latino female vs male) in Figure~\ref{fig: icd_distribution}.
While Overall male and female patients exhibit a similar trend in the leftist sub-figure, certain medical conditions show pronounced gender differences.
For example, cardiovascular diseases may be more prevalent among males~\cite{mercuro2010gender}, while autoimmune disorders are more common among females~\cite{quintero2012autoimmune}.

Furthermore, Figure~\ref{fig: icd_distribution} may reflect the distributional relation between overall and individual demographic groups. 
The majority group (White patients) closely mirrors the overall dataset's distribution, but minority groups display distinct patterns with different peaks.
These variations are critical because they can affect the fairness and accuracy of predictive models, which have been statistically analyzed in Section~\ref{sec:results}:
models trained predominantly on data from the majority group may not generalize well to minority populations with different disease prevalence patterns, potentially leading to biased predictions and exacerbating health inequalities.

\begin{table*}[htp]
\caption{Data percentage per group and performance table of GatorTron, Clinical BERT, and Clinical Longformer. Each value represents the mean and margin of error of the confidence interval. AA is short for African American, L is short for latino and NR is short for the race Not Reported.}
\label{tab: performance}
\resizebox{\textwidth}{!}{
\begin{tabular}{ll|l|cc|ccccc|ccccc|ccc}
 &  & All data & Male & Female & 18-29 & 30-49 & 50-69 & 70-89 & 90+ & White & Black/AA & Hispanic/L & Asian & Other/NR & Medicare & Medicaid & Other \\ \hline \hline
Data & \% & 100 & 48.72 & 51.28 & 7.45 & 20.11 & 40.05 & 29.06 & 3.33 & 68.99 & 14.68 & 5.15 & 3.15 & 8.04 & 41.18 & 7.84 & 50.98 \\ \hline

\multirow{6}{*}{GatorTron} & Acc & 10.36\textsubscript{ 0.13} & 9.35\textsubscript{ 0.09} & 11.32\textsubscript{ 0.18} & 27.47\textsubscript{ 0.52} & 18.13\textsubscript{ 0.49} & 8.64\textsubscript{ 0.12} & 4.82\textsubscript{ 0.01} & 3.19\textsubscript{ 0.55} & 9.63\textsubscript{ 0.28} & 10.04\textsubscript{ 0.68} & 13.48\textsubscript{ 0.08} & 16.70\textsubscript{ 1.10} & 12.72\textsubscript{ 0.11} & 4.84\textsubscript{ 0.18} & 12.18\textsubscript{ 0.11} & 14.54\textsubscript{ 0.14} \\
 & Precision & 65.59\textsubscript{ 0.18} & 65.58\textsubscript{ 0.22} & 65.59\textsubscript{ 0.14} & 34.54\textsubscript{ 0.18} & 52.70\textsubscript{ 0.11} & 68.75\textsubscript{ 0.20} & 75.13\textsubscript{ 0.30} & 76.32\textsubscript{ 0.65} & 66.40\textsubscript{ 0.14} & 66.61\textsubscript{ 0.62} & 59.54\textsubscript{ 0.15} & 58.80\textsubscript{ 0.13} & 63.35\textsubscript{ 0.26} & 72.80\textsubscript{ 0.39} & 58.50\textsubscript{ 0.36} & 60.85\textsubscript{ 0.09} \\
 & Recall & 46.88\textsubscript{ 0.65} & 46.75\textsubscript{ 0.65} & 46.99\textsubscript{ 0.64} & 22.89\textsubscript{ 0.49} & 36.75\textsubscript{ 0.76} & 48.49\textsubscript{ 0.75} & 55.39\textsubscript{ 0.68} & 55.56\textsubscript{ 0.74} & 47.65\textsubscript{ 0.58} & 47.82\textsubscript{ 0.78} & 43.29\textsubscript{ 1.07} & 41.29\textsubscript{ 1.32} & 43.02\textsubscript{ 0.44} & 53.15\textsubscript{ 0.62} & 40.15\textsubscript{ 0.33} & 42.85\textsubscript{ 0.72} \\
 & F1-mi & 60.36\textsubscript{ 0.35} & 60.30\textsubscript{ 0.34} & 60.42\textsubscript{ 0.37} & 44.30\textsubscript{ 0.03} & 54.99\textsubscript{ 0.46} & 59.65\textsubscript{ 0.48} & 63.07\textsubscript{ 0.23} & 63.41\textsubscript{ 0.28} & 60.46\textsubscript{ 0.33} & 60.97\textsubscript{ 0.45} & 60.26\textsubscript{ 0.25} & 58.72\textsubscript{ 0.64} & 58.74\textsubscript{ 0.37} & 62.06\textsubscript{ 0.27} & 56.21\textsubscript{ 0.43} & 58.85\textsubscript{ 0.44} \\
 & F1-ma & 46.19\textsubscript{ 0.13} & 45.33\textsubscript{ 0.19} & 44.87\textsubscript{ 0.22} & 33.91\textsubscript{ 2.76} & 44.83\textsubscript{ 0.48} & 45.71\textsubscript{ 0.13} & 45.50\textsubscript{ 0.51} & 43.79\textsubscript{ 1.06} & 45.77\textsubscript{ 0.23} & 46.37\textsubscript{ 0.03} & 46.14\textsubscript{ 0.28} & 45.56\textsubscript{ 0.69} & 46.09\textsubscript{ 0.29} & 46.05\textsubscript{ 0.31} & 45.07\textsubscript{ 0.21} & 46.04\textsubscript{ 0.06} \\
 & AUC & 94.17\textsubscript{ 0.14} & 94.00\textsubscript{ 0.17} & 94.33\textsubscript{ 0.12} & 95.16\textsubscript{ 0.00} & 94.82\textsubscript{ 0.10} & 93.92\textsubscript{ 0.14} & 93.57\textsubscript{ 0.19} & 93.37\textsubscript{ 0.16} & 94.09\textsubscript{ 0.15} & 94.28\textsubscript{ 0.07} & 94.72\textsubscript{ 0.17} & 94.45\textsubscript{ 0.06} & 94.17\textsubscript{ 0.17} & 93.59\textsubscript{ 0.18} & 94.24\textsubscript{ 0.19} & 94.55\textsubscript{ 0.10} \\ \hline
 
\multirow{6}{*}{\begin{tabular}[c]{@{}l@{}}Clinical \\ BERT\end{tabular}}
& Acc & 9.42\textsubscript{ 0.25} & 8.44\textsubscript{ 0.19} & 10.35\textsubscript{ 0.34} & 26.18\textsubscript{ 0.11} & 16.80\textsubscript{ 0.46} & 7.66\textsubscript{ 0.19} & 4.16\textsubscript{ 0.42} & 2.75\textsubscript{ 0.48} & 8.73\textsubscript{ 0.13} & 9.04\textsubscript{ 0.60} & 13.02\textsubscript{ 0.46} & 14.69\textsubscript{ 0.73} & 11.62\textsubscript{ 0.99} & 4.27\textsubscript{ 0.30} & 11.50\textsubscript{ 0.58} & 13.26\textsubscript{ 0.18} \\
 & Precision & 62.40\textsubscript{ 0.28} & 62.43\textsubscript{ 0.46} & 62.37\textsubscript{ 0.38} & 29.41\textsubscript{ 1.23} & 47.80\textsubscript{ 0.41} & 65.59\textsubscript{ 0.22} & 73.28\textsubscript{ 0.13} & 74.38\textsubscript{ 0.43} & 63.40\textsubscript{ 0.23} & 63.01\textsubscript{ 0.35} & 56.54\textsubscript{ 0.76} & 54.28\textsubscript{ 1.20} & 59.67\textsubscript{ 0.86} & 70.43\textsubscript{ 0.39} & 54.56\textsubscript{ 1.30} & 57.12\textsubscript{ 0.20} \\
 & Recall & 41.85\textsubscript{ 0.47} & 41.81\textsubscript{ 0.50} & 41.90\textsubscript{ 0.44} & 18.32\textsubscript{ 0.54} & 31.19\textsubscript{ 0.62} & 43.41\textsubscript{ 0.64} & 50.65\textsubscript{ 0.57} & 50.79\textsubscript{ 0.46} & 42.62\textsubscript{ 0.54} & 42.59\textsubscript{ 0.63} & 39.03\textsubscript{ 0.42} & 36.09\textsubscript{ 0.85} & 37.95\textsubscript{ 0.10} & 48.00\textsubscript{ 0.65} & 34.91\textsubscript{ 0.59} & 37.95\textsubscript{ 0.50} \\
 & F1-mi & 55.94\textsubscript{ 0.43} & 56.04\textsubscript{ 0.41} & 55.85\textsubscript{ 0.46} & 38.15\textsubscript{ 0.58} & 49.41\textsubscript{ 0.53} & 55.21\textsubscript{ 0.52} & 59.05\textsubscript{ 0.45} & 59.16\textsubscript{ 0.50} & 56.12\textsubscript{ 0.43} & 56.46\textsubscript{ 0.77} & 56.19\textsubscript{ 0.62} & 53.49\textsubscript{ 0.66} & 53.82\textsubscript{ 0.32} & 57.80\textsubscript{ 0.50} & 51.56\textsubscript{ 1.01} & 54.27\textsubscript{ 0.40} \\
 & F1-ma & 40.38\textsubscript{ 0.32} & 39.72\textsubscript{ 0.12} & 38.97\textsubscript{ 0.66} & 28.51\textsubscript{ 3.78} & 38.66\textsubscript{ 0.24} & 39.81\textsubscript{ 0.38} & 40.09\textsubscript{ 0.46} & 38.23\textsubscript{ 1.28} & 40.11\textsubscript{ 0.34} & 40.60\textsubscript{ 1.47} & 40.87\textsubscript{ 0.59} & 38.19\textsubscript{ 1.10} & 39.79\textsubscript{ 0.87} & 40.37\textsubscript{ 0.63} & 39.34\textsubscript{ 0.16} & 40.09\textsubscript{ 0.16} \\
 & AUC & 92.81\textsubscript{ 0.07} & 92.65\textsubscript{ 0.07} & 92.96\textsubscript{ 0.08} & 93.87\textsubscript{ 0.06} & 93.51\textsubscript{ 0.03} & 92.51\textsubscript{ 0.13} & 92.14\textsubscript{ 0.07} & 91.75\textsubscript{ 0.07} & 92.73\textsubscript{ 0.07} & 92.95\textsubscript{ 0.17} & 93.44\textsubscript{ 0.12} & 92.93\textsubscript{ 0.07} & 92.70\textsubscript{ 0.10} & 92.13\textsubscript{ 0.08} & 92.89\textsubscript{ 0.08} & 93.24\textsubscript{ 0.08} \\ \hline
 
 \multirow{6}{*}{\begin{tabular}[c]{@{}l@{}}Clinical \\ Longformer\end{tabular}}
& Acc & 12.69\textsubscript{ 0.42} & 11.52\textsubscript{ 0.53} & 13.81\textsubscript{ 0.33} & 30.91\textsubscript{ 1.01} & 21.08\textsubscript{ 0.62} & 10.73\textsubscript{ 0.46} & 6.85\textsubscript{ 0.73} & 5.32\textsubscript{ 0.51} & 11.94\textsubscript{ 0.54} & 12.54\textsubscript{ 0.30} & 16.38\textsubscript{ 0.51} & 19.44\textsubscript{ 1.45} & 14.38\textsubscript{ 0.82} & 6.96\textsubscript{ 0.38} & 15.10\textsubscript{ 0.20} & 16.96\textsubscript{ 0.50} \\
 & Precision & 70.10\textsubscript{ 0.79} & 70.44\textsubscript{ 0.83} & 69.77\textsubscript{ 0.76} & 44.58\textsubscript{ 1.21} & 59.48\textsubscript{ 0.57} & 72.69\textsubscript{ 0.83} & 77.95\textsubscript{ 1.29} & 79.04\textsubscript{ 0.74} & 70.77\textsubscript{ 0.90} & 70.92\textsubscript{ 0.90} & 65.78\textsubscript{ 0.23} & 65.03\textsubscript{ 0.27} & 67.59\textsubscript{ 0.49} & 76.12\textsubscript{ 1.17} & 64.35\textsubscript{ 0.53} & 66.11\textsubscript{ 0.64} \\
 & Recall & 60.39\textsubscript{ 3.39} & 60.13\textsubscript{ 3.42} & 60.64\textsubscript{ 3.37} & 36.52\textsubscript{ 2.96} & 50.02\textsubscript{ 3.12} & 61.85\textsubscript{ 3.48} & 69.05\textsubscript{ 3.48} & 70.55\textsubscript{ 4.09} & 61.07\textsubscript{ 3.31} & 61.55\textsubscript{ 3.77} & 57.63\textsubscript{ 3.25} & 55.71\textsubscript{ 2.78} & 56.06\textsubscript{ 3.77} & 67.16\textsubscript{ 3.59} & 54.33\textsubscript{ 3.92} & 55.86\textsubscript{ 3.15} \\
 & F1-mi & 70.77\textsubscript{ 1.80} & 70.51\textsubscript{ 1.82} & 71.05\textsubscript{ 1.78} & 59.18\textsubscript{ 2.41} & 66.29\textsubscript{ 1.81} & 69.97\textsubscript{ 1.84} & 73.12\textsubscript{ 1.71} & 74.06\textsubscript{ 2.00} & 70.80\textsubscript{ 1.72} & 71.41\textsubscript{ 1.92} & 71.03\textsubscript{ 2.23} & 70.50\textsubscript{ 1.61} & 69.17\textsubscript{ 2.15} & 72.24\textsubscript{ 1.76} & 67.55\textsubscript{ 1.96} & 69.44\textsubscript{ 1.83} \\
 & F1-ma & 60.52\textsubscript{ 4.75} & 59.80\textsubscript{ 4.58} & 59.19\textsubscript{ 4.75} & 49.19\textsubscript{ 5.22} & 58.97\textsubscript{ 4.12} & 60.01\textsubscript{ 4.77} & 60.23\textsubscript{ 5.01} & 57.41\textsubscript{ 5.65} & 60.25\textsubscript{ 4.75} & 60.36\textsubscript{ 4.70} & 60.54\textsubscript{ 5.43} & 59.76\textsubscript{ 4.86} & 59.53\textsubscript{ 5.05} & 60.38\textsubscript{ 5.04} & 58.70\textsubscript{ 4.43} & 60.46\textsubscript{ 4.52} \\
 & AUC & 96.58\textsubscript{ 0.51} & 96.48\textsubscript{ 0.53} & 96.68\textsubscript{ 0.50} & 96.99\textsubscript{ 0.47} & 96.94\textsubscript{ 0.42} & 96.44\textsubscript{ 0.55} & 96.27\textsubscript{ 0.54} & 96.13\textsubscript{ 0.79} & 96.54\textsubscript{ 0.52} & 96.69\textsubscript{ 0.48} & 96.99\textsubscript{ 0.46} & 96.71\textsubscript{ 0.61} & 96.38\textsubscript{ 0.50} & 96.27\textsubscript{ 0.56} & 96.64\textsubscript{ 0.51} & 96.77\textsubscript{ 0.48}
\\
\end{tabular}}
\end{table*}

\subsection{What Effects of the Imbalance Patterns will be for Model Performance and Fairness?}

The complex patterns of demographic imbalances identified in the dataset may have significant implications for clinical language models, which have rarely been explored in the existing studies. 
The skewed distributions of age, insurance coverage, and health condition documentation across gender and race suggest that models could inherit these biases if not appropriately addressed. 
Thus, a concrete question yet not been answered: What will the imbalance patterns impact model performance and fairness, especially for the demographic minority groups?
In the following sections, we conduct experiment and delve into the analysis of model performance and fairness across different demographic groups. 

\section{Experiments}

To evaluate the imbalance effects, we experimented clinical language models on the fundamental medical task, phenotype inference, predicting ICD-10 codes by discharge summary.
We include necessary demographic and SDoH attributes for the evaluation purpose and split the data 80/10/10 for the training, validation, and testing, respectively.
Our model choice depended on if a model achieved state-of-the-art performance and a model was pre-trained on clinical data. 
We included multiple state-of-the-art clinical language models (both generative and discriminative) for phenotype inference and finalized three models pretrained on MIMIC data, ClinicalBERT~\cite{alsentzer2019publicly}, GatorTron~\cite{yang2022gatortron}, and Clinical Longformer~\cite{li2022clinical}. 
Our experiments examined imbalance effects by broad performance metrics and included fairness by Equality Differences across diverse patient groups and their combinations, including 1) overall, genders, age ranges, ethnicity groups, and types of insurance, 2) the combinations of ethnicity and gender attributes (e.g., Asian - Male), and 3) combinations of insurance type and demographic value (e.g., Medicare - Hispanic).
We conducted further statistical analyses to provide more insights of relations between performance, fairness, and various imbalance patterns. 
We include implementation details in Appendix\ref{sub:appdix:implementation} to allow for reproduction.

\section{Results Analysis}
\label{sec:results}

\subsection{Performance Analysis}
\label{sec:perf}

After training each model that predicts ICD-10 code(s) (if any) of discharge summaries, we evaluated the three classification models on various metrics, including accuracy, precision, recall, F1 (-micro and -macro) scores, area under ROC curve (AUC).
We evaluate the model performance on the entire test dataset (overall performance) as well as across different subgroups based on gender, ethnicity or race, and insurance (a SDOH factor).
Table~\ref{tab: performance} presents the performance results, which indicate infer several critical findings:

\paragraph{Clinical Longformer outperforms Clinical BERT and GatorTron}
The Clinical Longformer outperforms the ClinicalBERT and GatorTron across almost all evaluation metrics by a large margin. 
For example, Clinical Longformer achieved a F1-micro score of 70.77\% on the entire test dataset, whereas GatorTron and ClinicalBERT only reached 60.36\%  and 55.94\%.
We infer that Clinical Longformer's ability to process a lengthier input allows it to capture more context and model dependencies of long clinical texts~\cite{li2023comparative}. 

\paragraph{Performance disparities among age groups}
A noteworthy pattern emerges when examining performance across different age groups.
Among the age groups, the youngest group (18-29 years) exhibits the greatest performance variance compared to other groups.
As shown in Table~\ref{tab: performance}, 
this group achieved the highest accuracy but scored lowest in all other metrics, including precision, recall, F1 scores, and AUC. 
Given that the young age group represents a smaller percentage of the MIMIC-IV (only 7.45\%), this may suggest that \textit{data imbalance negatively impacts the model's performance on the minor group}.
The divergence between accuracy and other metrics for the youngest age group is intriguing. 
This could be attributed to the model's tendency to predict fewer positive labels for this group, leading to high accuracy due to correct negative predictions, but lower precision and recall for the positive cases.

\paragraph{Performance disparities in race / ethnicity and insurance}
A similar trend is observed when analyzing race and ethnicity groups. 
The classifiers generally achieved higher performance on White and Black/African American patients across all metrics except accuracy, while relatively underperformed on other groups, such as Hispanic/Latino and Asian. 
Different insurance type subgroups also follow this same trend. 
Medicare and Other insurances count the majority of the data, and the model performances on Medicare and Other insurances are higher across all metrics except accuracy, than the Medicaid insurance group.
This disparity may reflect underlying data distribution imbalances or potential model biases caused by the data imbalance.

\paragraph{Data representation does not solely determine performance}
The relationship between data representation and model performance is not statistically correlated.
For instance, the 90+ age group, despite comprising only 3.33\% of the test data, achieved the highest performance across all metrics except accuracy. 
This suggests that raw data percentage alone does not determine model performance.
We infer that the eldest age group may share more similarities with the majority age groups (50-69 and 70-89) in terms of health conditions and corresponding diagnoses. 
This similarity in features (both in text encoding and label encoding) could explain the high performance despite low proportion.
From these observations, we infer that while data imbalance plays a role, feature similarity to the majority class may be a more critical factor in determining model performance. 
\textit{The model may overfit to majority data characteristics, leading to better performance on subgroups with similar features, regardless of their proportion in the dataset.}
To verify this hypothesis, we conducted statistical analyses in Section~\ref{subsec:stats_analy} to examine the correlation between data distances of subgroups and model performance.

\subsection{Fairness Analysis}
To assess the fairness of our models across different demographic groups, we employ the \emph{Equality Differences} (ED)~\cite{dixon2018measuring}. 
The equality difference quantifies the deviation of a group's performance from the overall performance and measures how equally and stably a model performs across various subgroups.
For each performance metric $m$ and a demographic group $g$ in $G$, the equality difference is defined as:
\begin{equation}
   ED_{g,m}=\sum_{g \in G}|P_{g,m}-P_{m}|
\label{eq: ED}
\end{equation}
Where $P_{g,m}$ is the performance of the model on group $g$ with respect to metric $m$, and $P_{m}$ is the model's performance on the entire test dataset for the same metric.
For example, the AUC equality difference is calculated by $\sum_{g\in G}|AUC_g-AUC|$, where $G$ is the gender and $g$ is a gender group (e.g., female).
A lower ED indicates that the model performs more similarly between the specific group and the overall population, suggesting greater fairness. 
The fairness evaluations over gender, age, race and insurance groups are in Table~\ref{tab: fairness}.

\begin{table}[htp]
\caption{equality difference over all metrics}
\label{tab: fairness}
\resizebox{0.485\textwidth}{!}{
\begin{tabular}{l|l|llllll}
Model & \% & Acc & P & R & F1-mi & F1-ma & AUC \\ \hline \hline
\multirow{4}{*}{GatorTron} & Gender & 1.97 & 0.01 & 0.24 & 0.12 & 2.18 & 0.33 \\
 & Age & 39.31 & 67.37 & 52.92 & 27.9 & 17.21 & 3.29 \\
 & Race & 12.87 & 16.91 & 14.75 & 4.07 & 1.38 & 1.02 \\
 & Insurance & 11.52 & 19.04 & 17.03 & 7.36 & 1.41 & 1.03 \\ \hline
\multirow{4}{*}{\begin{tabular}[c]{@{}l@{}}Clinical \\   BERT\end{tabular}} & Gender & 1.91 & 0.06 & 0.09 & 0.19 & 2.07 & 0.31 \\
 & Age & 37.83 & 73.64 & 53.49 & 31.38 & 16.6 & 3.79 \\
 & Race & 12.14 & 18.32 & 13.99 & 5.52 & 3.76 & 1.08 \\
 & Insurance & 11.07 & 21.15 & 16.99 & 7.91 & 1.34 & 1.19 \\ \hline
 
\multirow{4}{*}{\begin{tabular}[c]{@{}l@{}}Clinical \\  Longformer\end{tabular}} & Gender & 2.29 & 0.67 & 0.51 & 0.54 & 1.51 & 0.2 \\
 & Age & 41.78 & 55.52 & 54.52 & 22.51 & 15.44 & 1.67 \\
 & Race & 13.03 & 13.39 & 13.61 & 2.8 & 1.61 & 0.89 \\
 & Insurance & 12.41 & 15.76 & 17.36 & 6.02 & 1.89 & 0.56
\end{tabular}}
\end{table}

\paragraph{Unequal performance across age groups} The fairness evaluation in Table~\ref{tab: fairness} reinforces the disparities observed in model performance across age groups. The Equality Difference (ED) for age is significantly higher than that for gender, race, or insurance type, with values such as 37.83\% for accuracy and 73.64\% for precision in the Clinical BERT model. This substantial ED indicates that the models are less fair across different age groups, performing inconsistently and favoring certain age demographics over others. In contrast, the ED for gender is minimal (e.g., 1.91\% for accuracy in Clinical BERT), suggesting that the models are relatively fair across male and female groups.

\paragraph{Fairness across race and insurance types} While the ED values for race and insurance are lower than those for age, they are still notable. For race, the Clinical BERT model shows an ED of 12.14\% for accuracy and 18.32\% for precision, indicating some degree of performance variation among racial groups. Similarly, insurance types exhibit ED values such as 11.07\% for accuracy and 21.15\% for precision. These findings suggest that data imbalance across race and insurance subgroups can cause significant demographic disparities, a critical issue to be fixed.
Given the observed disparities in model performance across different demographic groups, it becomes essential to investigate the underlying factors contributing to these differences.
In the next section we conduct statistical analysis to uncover potential biases in data representation that may influence model fairness.

\begin{table*}[htp]
\centering
\caption{Cosine Distances between Each Group's Globally Average Label Vector.}
\label{tab: cosine_distance}
\begin{tabular}{ll|ll|lllll|p{0.6cm}p{0.6cm}p{0.7cm}p{0.6cm}p{0.6cm}|p{0.7cm}p{0.7cm}p{0.7cm}}
 & All & M & F & 18-29 & 30-49 & 50-69 &70-89 & 90+ & White &  Black &
  Hispanic &  Asian &  Other &  Medicare &  Medicaid &  Other \\ \hline \hline
All    & .000   & .014 & .015 & .357 & .095 & .007 & .026 & .092 & .002 & .026 & .029 & .039 & .009 & .008 & .050  & .007 \\ %
\hline 
Male            & .014 & .000  & .057 & .421 & .141 & .021 & .024 & .099 & .017 & .037 & .051 & .046 & .021 & .015 & .079 & .027 \\
Female          & .015 & .057 & .000  & .309 & .075 & .021 & .056 & .112 & .017 & .043 & .034 & .060  & .026 & .030  & .048 & .015 \\
\hline
18-29       & .357 & .421 & .309 & .000   & .128 & .353 & .508 & .569 & .361 & .363 & .310  & .456 & .366 & .427 & .213 & .310  \\
30-49       & .095 & .141 & .075 & .128 & .000  & .080  & .211 & .300   & .106 & .09  & .064 & .158 & .102 & .147 & .025 & .066 \\
50-69       & .007 & .021 & .021 & .353 & .080  & .000  & .049 & .140  & .012 & .025 & .023 & .046 & .012 & .024 & .044 & .004 \\
70-89       & .026 & .024 & .056 & .508 & .211 & .049 & .000   & .038 & .023 & .067 & .085 & .052 & .038 & .008 & .132 & .055 \\
90+         & .092 & .099 & .112 & .569 & .300   & .140  & .038 & .000   & .086 & .139 & .168 & .106 & .106 & .057 & .208 & .138 \\
\hline
White           & .002 & .017 & .017 & .361 & .106 & .012 & .023 & .086 & .000   & .042 & .043 & .051 & .014 & .008 & .06  & .011 \\
Black           & .026 & .037 & .043 & .363 & .090  & .025 & .067 & .139 & .042 & .000   & .017 & .043 & .034 & .039 & .05  & .030  \\ 
Hispanic & .029 & .051 & .034 & .310  & .064 & .023 & .085 & .168 & .043 & .017 & .000   & .051 & .034 & .050  & .034 & .025 \\ 
Asian           & .039 & .046 & .060  & .456 & .158 & .046 & .052 & .106 & .051 & .043 & .051 & .000   & .031 & .045 & .093 & .048 \\
Other & .009 & .021 & .026 & .366 & .102 & .012 & .038 & .106 & .014 & .034 & .034 & .031 & .000   & .022 & .054 & .011 \\
\hline
Medicare        & .008 & .015 & .030  & .427 & .147 & .024 & .008 & .057 & .008 & .039 & .050  & .045 & .022 & .000   & .085 & .029 \\
Medicaid        & .050  & .079 & .048 & .213 & .025 & .044 & .132 & .208 & .06  & .050  & .034 & .093 & .054 & .085 & .000   & .038 \\
Other & .007 & .027 & .015 & .310  & .066 & .004 & .055 & .138 & .011 & .030  & .025 & .048 & .011 & .029 & .038 & .000 
\end{tabular}%
\end{table*}

\subsection{Statistical Analysis}
\label{subsec:stats_analy}
Understanding the factors that influence model performance across different subgroups is essential for identifying potential biases and ensuring equitable outcomes. To this end, we conducted a statistical analysis to investigate whether the performance of our model is correlated with two possible factors: (a) the dissimilarity between a subgroup and the entire data, and (b) the proportion of each subgroup within the data.

Specifically, we sought to determine whether there is a statistically significant correlation between 
(a) the cosine distance between the globally averaged label vectors of each subgroup and that of the test data, and the performance of that subgroup, 
and (b) the proportion of the subgroup relative to the entire test data and its performance. 
We evaluated the correlation under the null hypothesis that there was no correlation between these factors, calculating the Pearson correlation coefficient and the p-value for each comparison.
\[H_0:\textrm{There is no correlation present between the factors. }\]
\[H_1: \textrm{There is a correlation present between the factors.} \]

Pearson Correlation Coefficient:
\[ r_i = \frac{\sum{(x_i-m_{x_i})(y-m_y)}}{\sqrt{\sum{(x_i-m_{x_i})^2}\sum{(y-m_y)^2}}}\]

In our analysis, \(x_i\) represents the vector of performance metrics (accuracy, precision, etc.) across various groups, with \(m_{x_i}\) being the mean of \(x_i\).
\(y\) denotes the vector of the values for which the metrics are being related (cosine distances in case (a) and proportion of subgroup in case (b)), and \(m_y\) is the mean of \(y\). In other words, the possible values of \(x_i\) and \(y\) are: \\ \indent $x = [\textrm{accuracy, precision, recall, F1-scores, roc-auc score}]$ \\ \indent $y = [\textrm{cosine distances, proportion of subgroup}]$

We used a significance level $\alpha$ of 0.05 to determine whether to reject the null hypothesis.
As previously determined, Clinical Longformer outperformed both GatorTron and ClinicalBERT during the performance evaluation and is able to capture more context, so we used Clinical Longformer for this correlation analysis, the results for which are presented in Table~\ref{tab: corr_distance_longfomer} and 
Table~\ref{tab: corr_proportion_longfomer}.

\begin{table*}[ht]
\centering
\caption{Correlation Analysis of Cosine Distance and Clinical Longformer Performance Metrics, correlation is flagged with an asterisk (*) when p value is less than 0.05}
\label{tab: corr_distance_longfomer}
\begin{tabular}{lcl|lcl|lcl}
\multicolumn{3}{c|}{\begin{tabular}[c]{@{}c@{}}Overall Data including all test data and each \\ individual gender, age range, race/ethnicity \\ group and type of insurance\end{tabular}} & \multicolumn{3}{c|}{Ethnicity/Race - Gender} & \multicolumn{3}{c}{Insurance - Ethnicity/Race} \\ \hline \hline
Metric & \multicolumn{1}{c}{Correlation} & P-Value & Metric & \multicolumn{1}{c}{Correlation} & P-Value & Metric & \multicolumn{1}{c}{Correlation} & P-Value \\
Accuracy & 0.713* & 1.95e-03* & Accuracy & 0.554 & 0.096 & Accuracy & 0.309 & 0.263 \\
Precision (Micro) & -0.815* & 1.21e-04* & Precision (Micro) & 0.144 & 0.692 & Precision (Micro) & -0.645* & 0.009* \\
Precision (Macro) & -0.950* & 1.94e-08* & Precision (Macro) & -0.360 & 0.307 & Precision (Macro) & -0.643* & 0.010* \\
Precision (Weighted) & -0.774* & 4.36e-04* & Precision (Weighted) & 0.305 & 0.392 & Precision (Weighted) & -0.402 & 0.137 \\
Precision (Samples) & -0.764* & 5.73e-04* & Precision (Samples) & -0.446 & 0.196 & Precision (Samples) & -0.423 & 0.116 \\
Recall (Micro) & -0.789* & 2.82e-04* & Recall (Micro) & 0.220 & 0.542 & Recall (Micro) & -0.414 & 0.125 \\
Recall (Macro) & -0.955* & 8.83E-09* & Recall (Macro) & -0.290 & 0.416 & Recall (Macro) & -0.619* & 0.014* \\
Recall (Weighted) & -0.789* & 2.82E-04* & Recall (Weighted) & 0.220 & 0.542 & Recall (Weighted) & -0.414 & 0.125 \\
Recall (Samples) & -0.715* & 1.86E-03* & Recall (Samples) & -0.340 & 0.337 & Recall (Samples) & -0.367 & 0.179 \\
F1 Score (Micro) & -0.806* & 1.61E-04* & F1 Score (Micro) & 0.264 & 0.461 & F1 Score (Micro) & -0.494 & 0.061 \\
F1 Score (Macro) & -0.973* & 2.76E-10* & F1 Score (Macro) & -0.606 & 0.063 & F1 Score (Macro) & -0.705* & 0.003* \\
F1 Score (Weighted) & -0.799* & 2.08E-04* & F1 Score (Weighted) & 0.200 & 0.579 & F1 Score (Weighted) & -0.467 & 0.079 \\
F1 Score (Samples) & -0.729* & 1.35E-03* & F1 Score (Samples) & -0.415 & 0.233 & F1 Score (Samples) & -0.400 & 0.140 \\
ROC-AUC (Micro) & -0.794* & 2.39E-04* & ROC-AUC (Micro) & 0.282 & 0.431 & ROC-AUC (Micro) & -0.411 & 0.128
\end{tabular}
\end{table*}

\begin{table*}[ht]
\centering
\caption{Correlation Analysis of Proportion of Subgroup and Clinical Longformer Performance Metrics}
\label{tab: corr_proportion_longfomer}
\begin{tabular}{lcc|lcc|lcc}
\multicolumn{3}{c|}{\begin{tabular}[c]{@{}c@{}}Overall Data including all test data and each \\ individual gender, age range, race/ethnicity \\ group and type of insurance\end{tabular}} & \multicolumn{3}{c|}{Ethnicity/Race - Gender} & \multicolumn{3}{c}{Insurance - Ethnicity/Race} \\ \hline \hline
Metric & \multicolumn{1}{c}{Correlation} & P-Value & Metric & \multicolumn{1}{c}{Correlation} & P-Value & Metric & \multicolumn{1}{c}{Correlation} & P-Value \\
Accuracy & -0.270 & 0.312 & Accuracy & -0.436 & 0.207 & Accuracy & -0.202 & 0.470 \\
Precision (Micro) & 0.175 & 0.518 & Precision (Micro) & -0.217 & 0.546 & Precision (Micro) & 0.157 & 0.577 \\
Precision (Macro) & 0.387 & 0.139 & Precision (Macro) & 0.340 & 0.336 & Precision (Macro) & 0.381 & 0.161 \\
Precision (Weighted) & 0.187 & 0.489 & Precision (Weighted) & 0.055 & 0.879 & Precision (Weighted) & 0.114 & 0.686 \\
Precision (Samples) & 0.267 & 0.318 & Precision (Samples) & 0.512 & 0.130 & Precision (Samples) & 0.218 & 0.434 \\
Recall (Micro) & 0.224 & 0.404 & Recall (Micro) & 0.224 & 0.534 & Recall (Micro) & 0.176 & 0.531 \\
Recall (Macro) & 0.341 & 0.196 & Recall (Macro) & 0.419 & 0.228 & Recall (Macro) & 0.261 & 0.348 \\
Recall (Weighted) & 0.224 & 0.404 & Recall (Weighted) & 0.224 & 0.534 & Recall (Weighted) & 0.176 & 0.531 \\
Recall (Samples) & 0.241 & 0.368 & Recall (Samples) & 0.607 & 0.062 & Recall (Samples) & 0.196 & 0.483 \\
F1 Score (Micro) & 0.216 & 0.423 & F1 Score (Micro) & 0.184 & 0.612 & F1 Score (Micro) & 0.181 & 0.520 \\
F1 Score (Macro) & 0.371 & 0.158 & F1 Score (Macro) & 0.557 & 0.095 & F1 Score (Macro) & 0.372 & 0.173 \\
F1 Score (Weighted) & 0.221 & 0.411 & F1 Score (Weighted) & 0.260 & 0.469 & F1 Score (Weighted) & 0.199 & 0.478 \\
F1 Score (Samples) & 0.247 & 0.356 & F1 Score (Samples) & 0.583 & 0.077 & F1 Score (Samples) & 0.205 & 0.463 \\
ROC-AUC (Micro) & 0.220 & 0.413 & ROC-AUC (Micro) & 0.168 & 0.643 & ROC-AUC (Micro) & 0.169 & 0.547
\end{tabular}
\end{table*}

\paragraph{\textbf{Performance of subgroups is more correlated with data dissimilarity, not data proportion}}
We observe significant correlations in Table~\ref{tab: corr_distance_longfomer} and no significant correlation in Table~\ref{tab: corr_proportion_longfomer}
This indicate that model performance can correlate with data dissimilarity (cosine distances) but not with the proportion of the subgroup within the test data. 
This result verifies our previous hypothesis that while data imbalance plays a role, \textit{feature similarity to the majority class may be a more critical factor in determining model performance}. 
The model may overfit to majority data characteristics, leading to better performance on subgroups with similar features, regardless of their proportion in the dataset.

\paragraph{\textbf{Accuracy is not a robust metric under data imbalance and long-tailed ICD-10 codes}}
Contrary to other metrics, we observed a positive correlation between accuracy and the cosine distances in the overall data, indicating that subgroups more dissimilar to the entire test data (higher cosine distance) tend to have higher accuracy.
This phenomenon may be due to how accuracy is calculated. 
Accuracy considers the number of predicted label vectors that exactly match the true label vector for a patient. 
Subgroups that are more different from the overall test data (e.g. young age group) may have fewer health issues or simpler ICD-10 code distributions, making it easier for the model to predict the exact label vector, thus resulting in higher accuracy. 
While other metrics are less strict and capture more nuanced information about individual ICD-10 code predictions. 
Therefore, when data has imbalance and long-tailed label distribution issues, accuracy may not be a robust metric and should be used with caution. 

\paragraph{\textbf{Macro scores are more stable under diverse imbalance correlation analysis}}
Although the overall data showed significant correlations across all metrics, the insurance-ethnicity/race subgroups showed significant correlations primarily in the macro-averaged metrics. 
We infer that as the impact of label imbalance within subgroups.
Macro-averaged metrics weigh each class equally and are more sensitive to class imbalances. 
In subgroups with fewer samples or uneven distributions of ICD-10 codes, the statistical power to identify significant correlations in other metrics is reduced. 
And, smaller subgroup sizes increase variability, making it harder to achieve statistical significance.
These may suggest that in subgroups with significant imbalance, metrics such as micro and samples scores may not reliably reflect true model performance (due to model's overfitting to majority).

\subsection{Case Studies}
To delve deeper into the implications of our statistical findings, we conducted a case study focusing on the combinations of insurance types and ethnicity/race. 
By examining these subgroups, we aimed to uncover potential factors contributing to the observed performance differences and provide deeper insights into the model's behavior across different demographic and the socioeconomic character.

We found that Clinical Longformer performed better overall for patients with Medicare and worse for patients with Medicaid. This trend was consistent across most race/ethnicity groups.
These results are similar to those in Table \ref{tab: performance}, which shows how well each model performed across different genders, age ranges, and other demographics, because all three models reported better scores for older patients—the main beneficiaries of Medicare.

\begin{figure}[ht]
\centering
    \includegraphics[width=.48\textwidth]{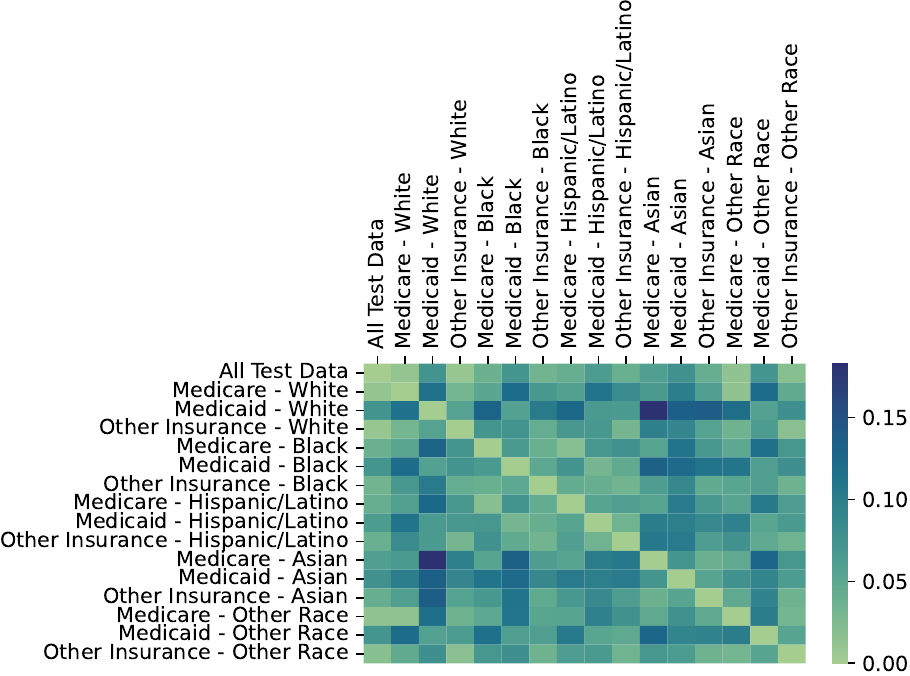}
    \caption{Cosine distances of label vectors between Insurance-Ethnicity.}
    \label{ins_race_cosine_distances_heatmap}
\end{figure}

Conversely, eligibility for Medicaid is based on income, not necessarily age, so it is possible that, due to the wider age range and varying circumstances, those with Medicaid may have less consistent label distributions. This may also explain the larger distance between the globally averaged label vectors of White individuals with Medicaid and Asian individuals with Medicare, as indicated by the heat map in Figure \ref{ins_race_cosine_distances_heatmap} and the corresponding distances in Table \ref{tab: ins_race_cosine_distance_table}. 

These observations highlight the importance of considering both demographic and socioeconomic factors when developing and evaluating models. 
Understanding how these factors influence model behavior can help in developing strategies to mitigate biases and enhance generalizability across diverse patient populations. 
Future work could involve adjusting for label imbalance or incorporating subgroup-specific training to improve performance where needed.

\begin{table*}
\caption{Performance table of Clinical BERT, GatorTron and Clinical Longformer for Insurance - Ethnicity/Race Subgroups. Mcare and Mcaid refer to Medicare and Medicaid, respectively. We use ``-W'', ``-B'', ``-A'', ``-H'', and ``-O'' to represent demographic groups of White American, Black/African American, Asian American, Hispanic/Latino American, and Others. Each performance value represents mean and margin of error from the confidence interval.}
\label{tab: performance-ins-race}
\centering
\resizebox{\textwidth}{!}{%
\begin{tabular}{ll|lllll|lllll|lllll}
 \multicolumn{2}{c}{Groups} & Mcare-W & Mcare-B & Mcare-H & Mcare-O & Mcare-A & Mcaid-W & Mcaid-B & Mcaid-H & Mcaid-O & Mcaid-A & Other-W & Other-B & Other-H & Other-O & Other-A\\ \hline \hline
Data & \% & 31.17 & 5.25 & 1.5 & 2.48 & 0.78 & 3.61 & 1.83 & 1.05 & 0.84 & 0.51 & 34.2 & 7.6 & 2.59 & 4.72 & 1.86 \\ \hline

\multirow{6}{*}{GatorTron} & Acc & 4.76\textsubscript{ 0.19} & 4.17\textsubscript{ 0.91} & 5.34\textsubscript{ 0.00} & 6.27\textsubscript{ 1.59} & 7.08\textsubscript{ 2.77} & 10.66\textsubscript{ 0.48} & 12.10\textsubscript{ 0.95} & 13.58\textsubscript{ 1.61} & 15.25\textsubscript{ 1.06} & 15.48\textsubscript{ 0.00} & 13.96\textsubscript{ 0.34} & 13.59\textsubscript{ 0.46} & 18.10\textsubscript{ 0.49} & 15.68\textsubscript{ 0.84} & 21.07\textsubscript{ 0.70} \\
 & Precision & 72.78\textsubscript{ 0.40} & 71.95\textsubscript{ 0.85} & 71.79\textsubscript{ 1.05} & 74.82\textsubscript{ 1.26} & 74.95\textsubscript{ 0.72} & 59.48\textsubscript{ 0.88} & 60.85\textsubscript{ 0.37} & 56.27\textsubscript{ 0.75} & 56.10\textsubscript{ 1.49} & 51.69\textsubscript{ 1.91} & 61.31\textsubscript{ 0.01} & 64.31\textsubscript{ 0.69} & 53.84\textsubscript{ 0.61} & 58.57\textsubscript{ 0.03} & 53.96\textsubscript{ 0.43} \\
 & Recall & 53.32\textsubscript{ 0.49} & 52.56\textsubscript{ 0.56} & 53.28\textsubscript{ 0.97} & 51.64\textsubscript{ 1.73} & 54.66\textsubscript{ 1.76} & 40.89\textsubscript{ 0.23} & 40.89\textsubscript{ 0.81} & 40.47\textsubscript{ 0.09} & 38.08\textsubscript{ 0.43} & 34.83\textsubscript{ 0.06} & 43.19\textsubscript{ 0.69} & 46.22\textsubscript{ 0.93} & 38.71\textsubscript{ 1.59} & 39.35\textsubscript{ 0.24} & 37.45\textsubscript{ 1.47} \\
 & F1-mi & 62.14\textsubscript{ 0.20} & 61.81\textsubscript{ 0.29} & 62.53\textsubscript{ 0.21} & 61.13\textsubscript{ 1.08} & 62.95\textsubscript{ 0.58} & 55.24\textsubscript{ 0.47} & 58.36\textsubscript{ 0.46} & 57.11\textsubscript{ 0.24} & 55.37\textsubscript{ 0.33} & 53.81\textsubscript{ 0.47} & 58.65\textsubscript{ 0.49} & 60.72\textsubscript{ 0.61} & 59.43\textsubscript{ 0.30} & 57.26\textsubscript{ 0.24} & 57.00\textsubscript{ 0.71} \\
 & F1-ma & 45.66\textsubscript{ 0.48} & 45.62\textsubscript{ 0.25} & 46.52\textsubscript{ 0.47} & 45.99\textsubscript{ 1.38} & 45.67\textsubscript{ 1.01} & 43.48\textsubscript{ 0.40} & 45.39\textsubscript{ 0.62} & 41.33\textsubscript{ 0.15} & 43.49\textsubscript{ 0.48} & 39.20\textsubscript{ 0.62} & 45.48\textsubscript{ 0.14} & 46.86\textsubscript{ 0.15} & 46.64\textsubscript{ 0.87} & 45.30\textsubscript{ 0.29} & 42.84\textsubscript{ 0.29} \\
 & AUC & 93.55\textsubscript{ 0.19} & 93.67\textsubscript{ 0.15} & 94.35\textsubscript{ 0.27} & 93.42\textsubscript{ 0.10} & 93.60\textsubscript{ 0.09} & 94.14\textsubscript{ 0.28} & 94.64\textsubscript{ 0.01} & 94.25\textsubscript{ 0.12} & 94.05\textsubscript{ 0.30} & 93.61\textsubscript{ 0.33} & 94.49\textsubscript{ 0.11} & 94.54\textsubscript{ 0.04} & 94.99\textsubscript{ 0.12} & 94.49\textsubscript{ 0.19} & 94.93\textsubscript{ 0.01} \\    \hline

\multirow{6}{*}{\begin{tabular}[c]{@{}l@{}}Clinical \\ BERT\end{tabular}}
& Acc & 4.22\textsubscript{ 0.26} & 3.92\textsubscript{ 0.46} & 5.27\textsubscript{ 1.02} & 4.46\textsubscript{ 0.47} & 5.92\textsubscript{ 2.00} & 10.10\textsubscript{ 0.55} & 11.11\textsubscript{ 2.26} & 13.48\textsubscript{ 0.70} & 14.02\textsubscript{ 1.59} & 14.68\textsubscript{ 1.71} & 12.69\textsubscript{ 0.12} & 12.07\textsubscript{ 0.32} & 17.27\textsubscript{ 0.43} & 14.97\textsubscript{ 1.22} & 18.37\textsubscript{ 0.23} \\
 & Precision & 70.60\textsubscript{ 0.33} & 68.59\textsubscript{ 0.12} & 69.39\textsubscript{ 2.75} & 72.73\textsubscript{ 0.29} & 70.83\textsubscript{ 2.61} & 56.04\textsubscript{ 0.49} & 56.48\textsubscript{ 1.47} & 53.35\textsubscript{ 2.21} & 51.01\textsubscript{ 3.94} & 45.41\textsubscript{ 4.58} & 57.62\textsubscript{ 0.32} & 60.73\textsubscript{ 0.41} & 50.48\textsubscript{ 0.59} & 54.31\textsubscript{ 0.97} & 49.75\textsubscript{ 0.57} \\
 & Recall & 48.20\textsubscript{ 0.57} & 47.25\textsubscript{ 1.40} & 48.86\textsubscript{ 0.20} & 46.59\textsubscript{ 0.97} & 47.80\textsubscript{ 0.90} & 35.44\textsubscript{ 0.68} & 35.39\textsubscript{ 1.24} & 36.70\textsubscript{ 0.52} & 31.97\textsubscript{ 2.42} & 30.27\textsubscript{ 2.46} & 38.30\textsubscript{ 0.57} & 41.10\textsubscript{ 0.37} & 34.34\textsubscript{ 0.96} & 34.44\textsubscript{ 0.06} & 32.76\textsubscript{ 0.54} \\
 & F1-mi & 57.92\textsubscript{ 0.44} & 57.40\textsubscript{ 1.12} & 59.07\textsubscript{ 0.56} & 56.64\textsubscript{ 0.11} & 57.08\textsubscript{ 0.96} & 50.59\textsubscript{ 0.89} & 53.24\textsubscript{ 1.60} & 54.29\textsubscript{ 0.75} & 49.12\textsubscript{ 1.94} & 49.29\textsubscript{ 1.93} & 54.17\textsubscript{ 0.45} & 56.24\textsubscript{ 0.45} & 54.15\textsubscript{ 1.10} & 52.21\textsubscript{ 0.44} & 52.06\textsubscript{ 0.45} \\
 & F1-ma & 40.08\textsubscript{ 0.53} & 39.57\textsubscript{ 1.01} & 41.30\textsubscript{ 1.46} & 40.38\textsubscript{ 0.83} & 37.69\textsubscript{ 1.96} & 38.41\textsubscript{ 0.82} & 38.67\textsubscript{ 1.70} & 37.84\textsubscript{ 1.87} & 36.67\textsubscript{ 2.06} & 34.14\textsubscript{ 3.19} & 39.74\textsubscript{ 0.31} & 41.11\textsubscript{ 1.25} & 40.35\textsubscript{ 0.39} & 38.83\textsubscript{ 1.21} & 36.05\textsubscript{ 1.43} \\
 & AUC & 92.12\textsubscript{ 0.08} & 92.16\textsubscript{ 0.17} & 92.68\textsubscript{ 0.19} & 91.88\textsubscript{ 0.17} & 91.90\textsubscript{ 0.07} & 92.74\textsubscript{ 0.09} & 93.28\textsubscript{ 0.29} & 93.30\textsubscript{ 0.15} & 92.37\textsubscript{ 0.25} & 92.36\textsubscript{ 0.71} & 93.18\textsubscript{ 0.07} & 93.34\textsubscript{ 0.15} & 93.80\textsubscript{ 0.12} & 93.07\textsubscript{ 0.05} & 93.41\textsubscript{ 0.10} \\ \hline
 
 \multirow{6}{*}{\begin{tabular}[c]{@{}l@{}}Clinical \\ Longformer\end{tabular}} & Acc & 6.90\textsubscript{ 0.40} & 7.01\textsubscript{ 0.43} & 7.38\textsubscript{ 2.47} & 5.77\textsubscript{ 2.14} & 11.84\textsubscript{ 2.77} & 13.24\textsubscript{ 1.68} & 15.79\textsubscript{ 0.95} & 16.20\textsubscript{ 1.76} & 18.20\textsubscript{ 1.40} & 18.45\textsubscript{ 5.33} & 16.40\textsubscript{ 0.59} & 15.59\textsubscript{ 0.26} & 21.61\textsubscript{ 1.49} & 18.25\textsubscript{ 0.94} & 22.91\textsubscript{ 1.99} \\
 & Precision & 76.08\textsubscript{ 1.23} & 76.12\textsubscript{ 1.31} & 75.27\textsubscript{ 0.50} & 76.51\textsubscript{ 0.95} & 78.24\textsubscript{ 1.77} & 65.55\textsubscript{ 1.37} & 65.40\textsubscript{ 1.39} & 63.93\textsubscript{ 0.24} & 60.61\textsubscript{ 2.38} & 58.86\textsubscript{ 2.36} & 66.48\textsubscript{ 0.74} & 68.66\textsubscript{ 0.66} & 61.10\textsubscript{ 0.61} & 64.12\textsubscript{ 0.47} & 61.17\textsubscript{ 0.24} \\
 & Recall & 67.23\textsubscript{ 3.54} & 67.32\textsubscript{ 3.51} & 67.73\textsubscript{ 4.96} & 64.96\textsubscript{ 3.74} & 69.11\textsubscript{ 3.30} & 54.98\textsubscript{ 4.63} & 54.50\textsubscript{ 3.39} & 56.53\textsubscript{ 1.87} & 50.98\textsubscript{ 5.85} & 49.80\textsubscript{ 2.79} & 56.10\textsubscript{ 2.97} & 59.27\textsubscript{ 4.06} & 52.29\textsubscript{ 2.84} & 52.26\textsubscript{ 3.52} & 51.69\textsubscript{ 2.69} \\
 & F1-mi & 72.27\textsubscript{ 1.75} & 72.37\textsubscript{ 1.56} & 72.56\textsubscript{ 2.72} & 70.99\textsubscript{ 1.92} & 73.50\textsubscript{ 1.62} & 67.16\textsubscript{ 2.26} & 68.15\textsubscript{ 1.89} & 69.38\textsubscript{ 1.49} & 65.77\textsubscript{ 2.81} & 66.50\textsubscript{ 1.43} & 69.13\textsubscript{ 1.63} & 71.17\textsubscript{ 2.30} & 70.28\textsubscript{ 2.20} & 68.20\textsubscript{ 2.27} & 69.47\textsubscript{ 2.08} \\
 & F1-ma & 60.14\textsubscript{ 5.13} & 60.11\textsubscript{ 4.40} & 59.80\textsubscript{ 5.62} & 58.82\textsubscript{ 5.28} & 59.14\textsubscript{ 5.35} & 57.72\textsubscript{ 4.49} & 57.39\textsubscript{ 3.82} & 57.31\textsubscript{ 5.41} & 55.50\textsubscript{ 5.51} & 51.92\textsubscript{ 2.56} & 60.03\textsubscript{ 4.31} & 60.62\textsubscript{ 5.01} & 60.91\textsubscript{ 5.83} & 59.74\textsubscript{ 4.86} & 58.03\textsubscript{ 5.23} \\
 & AUC & 96.27\textsubscript{ 0.56} & 96.33\textsubscript{ 0.56} & 96.67\textsubscript{ 0.59} & 95.94\textsubscript{ 0.48} & 96.17\textsubscript{ 0.86} & 96.63\textsubscript{ 0.53} & 96.84\textsubscript{ 0.50} & 96.75\textsubscript{ 0.58} & 96.08\textsubscript{ 0.37} & 96.53\textsubscript{ 0.58} & 96.73\textsubscript{ 0.49} & 96.86\textsubscript{ 0.42} & 97.19\textsubscript{ 0.35} & 96.61\textsubscript{ 0.55} & 96.93\textsubscript{ 0.54}\\
\end{tabular}}
\end{table*}

\begin{table*}[t]
\centering
\caption{Cosine Distances between Each Group's Globally Average Label Vector for Insurance - Ethnicity/Race Subgroups. We bolden the largest and underline the smallest value in each row.}
\resizebox{\textwidth}{!}{
\label{tab: ins_race_cosine_distance_table}
\begin{tabular}{p{1.2cm}l|l|p{0.7cm}p{0.7cm}p{0.7cm}|p{0.7cm}p{0.7cm}p{0.7cm}|p{0.7cm}p{0.7cm}p{0.7cm}|p{0.7cm}p{0.7cm}p{0.7cm}|p{0.7cm}p{0.7cm}p{0.7cm}}
 &   & & \multicolumn{3}{c|}{White} &  \multicolumn{3}{c|}{Black/African American} &  \multicolumn{3}{c|}{Hispanic/Latino} &  \multicolumn{3}{c|}{Asian} &  \multicolumn{3}{c}{Other/Not Reported} \\ &   &  All  &  Medicare &  Medicaid &  Other &  Medicare &  Medicaid &  Other &  Medicare &  Medicaid &  Other &  Medicare &  Medicaid &  Other &  Medicare &  Medicaid &  Other \\ \hline \hline
 &  All Test Data &  .000 &  .011 &  .074 &  {\ul .009} &  .038 &  .072 &  .033 &  .041 &  .064 &  .040 &  .060 &  \textbf{.077} &  .041 &  .014 &  .073 &  .019 \\ \hline
 &  Medicare &  {\ul .011} &  .000 &  .116 &  .032 &  .051 &  \textbf{.122} &  .069 &  .059 &  .112 &  .085 &  .067 &  .100 &  .061 &  .013 &  \textbf{.122} &  .046 \\
 &  Medicaid &  .074 &  .116 &  .000 &  {\ul .056} &  .131 &  .058 &  .103 &  .126 &  .067 &  .065 &  \textbf{.183} &  .135 &  .138 &  .119 &  .058 &  .080 \\
\multirow{-3}{*}{White} &  Other &  {\ul .009} &  .032 &  .056 &  .000 &  .072 &  .074 &  .043 &  .070 &  .069 &  .031 &  \textbf{.098} &  .091 &  .057 &  .036 &  .064 &  .018 \\ \hline
 &  Medicare &  .038 &  .051 &  \textbf{.131} &  .072 &  .000 &  .067 &  .040 &  {\ul .019} &  .070 &  .077 &  .055 &  .112 &  .069 &  .049 &  .117 &  .070 \\
 &  Medicaid &  .072 &  .122 &  .058 &  .074 &  .067 &  .000 &  .049 &  .075 &  {\ul .031} &  .047 &  \textbf{.134} &  .123 &  .112 &  .111 &  .061 &  .080 \\
\multirow{-3}{*}{\begin{tabular}[c]{@{}l@{}}Black/\\ African\\ American\end{tabular}} &  Other &  {\ul .033} &  .069 &  \textbf{.103} &  .043 &  .040 &  .049 &  .000 &  .044 &  .041 &  {\ul .033} &  .063 &  .088 &  .047 &  .053 &  .062 &  .036 \\ \hline
 &  Medicare &  .041 &  .059 &  \textbf{.126} &  .070 &  {\ul .019} &  .075 &  .044 &  .000 &  .055 &  .060 &  .055 &  .105 &  .068 &  .055 &  .106 &  .063 \\
 &  Medicaid &  .064 &  \textbf{.112} &  .067 &  .069 &  .070 &  {\ul .031} &  .041 &  .055 &  .000 &  .035 &  .101 &  .099 &  .087 &  .096 &  .053 &  .067 \\
\multirow{-3}{*}{\begin{tabular}[c]{@{}l@{}}Hispanic/\\ Latino\end{tabular}} &  Other &  .040 &  .085 &  .065 &  {\ul .031} &  .077 &  .047 &  .033 &  .060 &  .035 &  .000 &  \textbf{.107} &  .105 &  .063 &  .077 &  .048 &  .035 \\ \hline
 &  Medicare &  .060 &  .067 &  \textbf{.183} &  .098 &  .055 &  .134 &  .063 &  .055 &  .101 &  .107 &  .000 &  .072 &  {\ul .038} &  .048 &  .127 &  .070 \\
 &  Medicaid &  .077 &  .100 &  \textbf{.135} &  .091 &  .112 &  .123 &  .088 &  .105 &  .099 &  .105 &  .072 &  .000 &  {\ul .055} &  .077 &  .093 &  .064 \\
\multirow{-3}{*}{Asian} &  Other &  .041 &  .061 &  \textbf{.138} &  .057 &  .069 &  .112 &  .047 &  .068 &  .087 &  .063 &  .038 &  .055 &  .000 &  .048 &  .094 &  {\ul .036} \\ \hline
 &  Medicare &  .014 &  {\ul .013} &  \textbf{.119} &  .036 &  .049 &  .111 &  .053 &  .055 &  .096 &  .077 &  .048 &  .077 &  .048 &  .000 &  .101 &  .032 \\
 &  Medicaid &  .073 &  .122 &  .058 &  .064 &  .117 &  .061 &  .062 &  .106 &  .053 &  {\ul .048} &  \textbf{.127} &  .093 &  .094 &  .101 &  .000 &  .054 \\
\multirow{-3}{*}{\begin{tabular}[c]{@{}l@{}}Other/\\ Not \\Reported\end{tabular}} &  Other &  .019 &  .046 &  \textbf{.080} &  {\ul .018} &  .070 &  \textbf{.080} &  .036 &  .063 &  .067 &  .035 &  .070 &  .064 &  .036 &  .032 &  .054 &  .000
\end{tabular}%
}
\end{table*}

\section{Conclusion and Limitations}
In this study, we explored how data imbalance impacts the ICD-10 code predictions of state-of-the-art clinical language models while examining this imbalance and scrutinizing their performance and fairness for specific groups based on age, gender, ethnic group, and type of insurance. We also investigated the various subgroups of these characteristics. In doing so, we found that that the highest variance in performance was for the age groups, with further analysis revealing that these disparities were closely tied to the dissimilarity between subgroup data and the overall test data. 
The models had more trouble predicting the ICD-10 codes of patients that belonged to groups that were most dissimilar to test data distributions, regardless of the subgroup’s representation within the test data. 
In addition, the case study on ethnicity and insurance revealed that the model performed better for patients with Medicare but worse for those with Medicaid, suggesting socioeconomic factors may play a role in these disparities.

While we have examined imbalance effects and performance, two major limitations have be acknowledged to appropriately interpret our findings. First, we only experimented the phenotype inference on the MIMIC-IV, limiting the generalizability of our findings to other clinical LLMs, other clinical tasks, and non-ICU databases. 
Second, cosine distance was the metric for data dissimilarity, and alternative metrics may measure different ICD-10 code distribution distances.

In conclusion, our study highlights the critical need for further investigation into the fairness and performance of clinical large language models, particularly in relation to underrepresented groups. Addressing these challenges will be essential to ensure more equitable diagnostic predictions.

\section*{Acknowledgment}
The authors thank anonymous reviewers for their insightful feedback. 
This project was supported by the National Science Foundation (NSF) IIS-2245920 and National Institutes of Health (NIH) National Cancer Institute under award R01 CA202210 and R01 CA258193.
We thank for the computing resources provided by the NAIRR-Pilot (NAIRR240165).

\bibliographystyle{IEEEtran}  
\bibliography{reference}

\begin{thebibliography}{10}
\providecommand{\url}[1]{#1}
\csname url@samestyle\endcsname
\providecommand{\newblock}{\relax}
\providecommand{\bibinfo}[2]{#2}
\providecommand{\BIBentrySTDinterwordspacing}{\spaceskip=0pt\relax}
\providecommand{\BIBentryALTinterwordstretchfactor}{4}
\providecommand{\BIBentryALTinterwordspacing}{\spaceskip=\fontdimen2\font plus
\BIBentryALTinterwordstretchfactor\fontdimen3\font minus \fontdimen4\font\relax}
\providecommand{\BIBforeignlanguage}[2]{{%
\expandafter\ifx\csname l@#1\endcsname\relax
\typeout{** WARNING: IEEEtran.bst: No hyphenation pattern has been}%
\typeout{** loaded for the language `#1'. Using the pattern for}%
\typeout{** the default language instead.}%
\else
\language=\csname l@#1\endcsname
\fi
#2}}
\providecommand{\BIBdecl}{\relax}
\BIBdecl

\bibitem{wu2023token}
\BIBentryALTinterwordspacing
Y.~Wu, I.-C. Huang, and X.~Huang, ``Token imbalance adaptation for radiology report generation,'' in \emph{Proceedings of the Conference on Health, Inference, and Learning}, ser. Proceedings of Machine Learning Research, B.~J. Mortazavi, T.~Sarker, A.~Beam, and J.~C. Ho, Eds., vol. 209.\hskip 1em plus 0.5em minus 0.4em\relax PMLR, 22 Jun--24 Jun 2023, pp. 72--85. [Online]. Available: \url{https://proceedings.mlr.press/v209/wu23a.html}
\BIBentrySTDinterwordspacing

\bibitem{johnson2023mimic}
\BIBentryALTinterwordspacing
A.~E.~W. Johnson, L.~Bulgarelli, L.~Shen, A.~Gayles, A.~Shammout, S.~Horng, T.~J. Pollard, S.~Hao, B.~Moody, B.~Gow, L.-w.~H. Lehman, L.~A. Celi, and R.~G. Mark, ``{MIMIC-IV, a freely accessible electronic health record dataset},'' \emph{Scientific Data}, vol.~10, no.~1, p.~1, 2023. [Online]. Available: \url{https://doi.org/10.1038/s41597-022-01899-x}
\BIBentrySTDinterwordspacing

\bibitem{cloutier2023fine}
N.~A. Cloutier and N.~Japkowicz, ``Fine-tuned generative llm oversampling can improve performance over traditional techniques on multiclass imbalanced text classification,'' in \emph{2023 IEEE International Conference on Big Data (BigData)}.\hskip 1em plus 0.5em minus 0.4em\relax IEEE, 2023, pp. 5181--5186.

\bibitem{wornow2023shaky}
\BIBentryALTinterwordspacing
M.~Wornow, Y.~Xu, R.~Thapa, B.~Patel, E.~Steinberg, S.~Fleming, M.~A. Pfeffer, J.~Fries, and N.~H. Shah, ``{The shaky foundations of large language models and foundation models for electronic health records},'' \emph{npj Digital Medicine}, vol.~6, no.~1, p. 135, jul 2023. [Online]. Available: \url{https://www.nature.com/articles/s41746-023-00879-8}
\BIBentrySTDinterwordspacing

\bibitem{wang2020imbalanced}
S.~Wang, M.~E. Elkin, and X.~Zhu, ``Imbalanced learning for hospital readmission prediction using national readmission database,'' in \emph{2020 IEEE International Conference on Knowledge Graph (ICKG)}.\hskip 1em plus 0.5em minus 0.4em\relax IEEE, 2020, pp. 116--122.

\bibitem{lu2022comparative}
H.~Lu, L.~Ehwerhemuepha, and C.~Rakovski, ``A comparative study on deep learning models for text classification of unstructured medical notes with various levels of class imbalance,'' \emph{BMC medical research methodology}, vol.~22, no.~1, p. 181, 2022.

\bibitem{angeli2022class}
\BIBentryALTinterwordspacing
K.~{De Angeli}, S.~Gao, I.~Danciu, E.~B. Durbin, X.-C. Wu, A.~Stroup, J.~Doherty, S.~Schwartz, C.~Wiggins, M.~Damesyn, L.~Coyle, L.~Penberthy, G.~D. Tourassi, and H.-J. Yoon, ``Class imbalance in out-of-distribution datasets: Improving the robustness of the textcnn for the classification of rare cancer types,'' \emph{Journal of Biomedical Informatics}, vol. 125, p. 103957, 2022. [Online]. Available: \url{https://www.sciencedirect.com/science/article/pii/S1532046421002860}
\BIBentrySTDinterwordspacing

\bibitem{henning2023survey}
\BIBentryALTinterwordspacing
S.~Henning, W.~Beluch, A.~Fraser, and A.~Friedrich, ``A survey of methods for addressing class imbalance in deep-learning based natural language processing,'' in \emph{Proceedings of the 17th Conference of the European Chapter of the Association for Computational Linguistics}, A.~Vlachos and I.~Augenstein, Eds.\hskip 1em plus 0.5em minus 0.4em\relax Dubrovnik, Croatia: Association for Computational Linguistics, May 2023, pp. 523--540. [Online]. Available: \url{https://aclanthology.org/2023.eacl-main.38}
\BIBentrySTDinterwordspacing

\bibitem{kino2021scoping}
\BIBentryALTinterwordspacing
S.~Kino, Y.-T. Hsu, K.~Shiba, Y.-S. Chien, C.~Mita, I.~Kawachi, and A.~Daoud, ``{A scoping review on the use of machine learning in research on social determinants of health: Trends and research prospects},'' \emph{SSM - Population Health}, vol.~15, p. 100836, 2021. [Online]. Available: \url{https://www.sciencedirect.com/science/article/pii/S2352827321001117}
\BIBentrySTDinterwordspacing

\bibitem{roosli2022peeking}
\BIBentryALTinterwordspacing
E.~R{\"{o}}{\"{o}}sli, S.~Bozkurt, and T.~Hernandez-Boussard, ``{Peeking into a black box, the fairness and generalizability of a MIMIC-III benchmarking model},'' \emph{Scientific Data}, vol.~9, no.~1, p.~24, jan 2022. [Online]. Available: \url{https://www.nature.com/articles/s41597-021-01110-7}
\BIBentrySTDinterwordspacing

\bibitem{yang2023evaluating}
\BIBentryALTinterwordspacing
M.~Y. Yang, G.~H. Kwak, T.~Pollard, L.~A. Celi, and M.~Ghassemi, ``Evaluating the impact of social determinants on health prediction in the intensive care unit,'' in \emph{Proceedings of the 2023 AAAI/ACM Conference on AI, Ethics, and Society}, ser. AIES '23.\hskip 1em plus 0.5em minus 0.4em\relax New York, NY, USA: Association for Computing Machinery, 2023, p. 333–350. [Online]. Available: \url{https://doi.org/10.1145/3600211.3604719}
\BIBentrySTDinterwordspacing

\bibitem{johnson2023mimic-note}
\BIBentryALTinterwordspacing
A.~Johnson, T.~Pollard, S.~Horng, L.~A. Celi, and R.~Mark, ``Mimic-iv-note: Deidentified free-text clinical notes (version 2.2),'' 2023. [Online]. Available: \url{https://doi.org/10.13026/1n74-ne17}
\BIBentrySTDinterwordspacing

\bibitem{alsentzer2019publicly}
\BIBentryALTinterwordspacing
E.~Alsentzer, J.~Murphy, W.~Boag, W.-H. Weng, D.~Jindi, T.~Naumann, and M.~McDermott, ``Publicly available clinical {BERT} embeddings,'' in \emph{Proceedings of the 2nd Clinical Natural Language Processing Workshop}, A.~Rumshisky, K.~Roberts, S.~Bethard, and T.~Naumann, Eds.\hskip 1em plus 0.5em minus 0.4em\relax Minneapolis, Minnesota, USA: Association for Computational Linguistics, Jun. 2019, pp. 72--78. [Online]. Available: \url{https://aclanthology.org/W19-1909}
\BIBentrySTDinterwordspacing

\bibitem{yang2022gatortron}
\BIBentryALTinterwordspacing
X.~Yang, A.~Chen, N.~PourNejatian, H.~C. Shin, K.~E. Smith, C.~Parisien, C.~Compas, C.~Martin, M.~G. Flores, Y.~Zhang, T.~Magoc, C.~A. Harle, G.~Lipori, D.~A. Mitchell, W.~R. Hogan, E.~A. Shenkman, J.~Bian, and Y.~Wu, ``{GatorTron: A Large Clinical Language Model to Unlock Patient Information from Unstructured Electronic Health Records},'' \emph{arXiv preprint arXiv:2203.03540}, feb 2022. [Online]. Available: \url{http://arxiv.org/abs/2203.03540}
\BIBentrySTDinterwordspacing

\bibitem{li2022clinical}
Y.~Li, R.~M. Wehbe, F.~S. Ahmad, H.~Wang, and Y.~Luo, ``Clinical-longformer and clinical-bigbird: Transformers for long clinical sequences,'' 2022.

\bibitem{biseda2020predictionicdcodesclinical}
\BIBentryALTinterwordspacing
B.~Biseda, G.~Desai, H.~Lin, and A.~Philip, ``Prediction of icd codes with clinical bert embeddings and text augmentation with label balancing using mimic-iii,'' 2020. [Online]. Available: \url{https://arxiv.org/abs/2008.10492}
\BIBentrySTDinterwordspacing

\bibitem{ji2021does}
S.~Ji, M.~H{\"o}ltt{\"a}, and P.~Marttinen, ``Does the magic of bert apply to medical code assignment? a quantitative study,'' \emph{Computers in biology and medicine}, vol. 139, p. 104998, 2021.

\bibitem{tsai2019leveraging}
S.-C. Tsai, T.-Y. Chang, and Y.-N. Chen, ``Leveraging hierarchical category knowledge for data-imbalanced multi-label diagnostic text understanding,'' in \emph{Proceedings of the 10th international workshop on health text mining and information analysis (LOUHI)}, 2019, pp. 39--43.

\bibitem{pmlr-v106-xu19a}
\BIBentryALTinterwordspacing
K.~Xu, M.~Lam, J.~Pang, X.~Gao, C.~Band, P.~Mathur, F.~Papay, A.~K. Khanna, J.~B. Cywinski, K.~Maheshwari, P.~Xie, and E.~P. Xing, ``Multimodal machine learning for automated icd coding,'' in \emph{Proceedings of the 4th Machine Learning for Healthcare Conference}, ser. Proceedings of Machine Learning Research, vol. 106.\hskip 1em plus 0.5em minus 0.4em\relax PMLR, 09--10 Aug 2019, pp. 197--215. [Online]. Available: \url{https://proceedings.mlr.press/v106/xu19a.html}
\BIBentrySTDinterwordspacing

\bibitem{jin2023learning}
Y.~Jin, Y.~Xiong, D.~Shi, Y.~Lin, L.~He, Y.~Zhang, J.~M. Plasek, L.~Zhou, D.~W. Bates, and C.~Tang, ``Learning from undercoded clinical records for automated international classification of diseases (icd) coding,'' \emph{Journal of the American Medical Informatics Association}, vol.~30, no.~3, pp. 438--446, 2023.

\bibitem{nerella2024transformers}
S.~Nerella, S.~Bandyopadhyay, J.~Zhang, M.~Contreras, S.~Siegel, A.~Bumin, B.~Silva, J.~Sena, B.~Shickel, A.~Bihorac \emph{et~al.}, ``Transformers and large language models in healthcare: A review,'' \emph{Artificial Intelligence in Medicine}, p. 102900, 2024.

\bibitem{kuroiwa2023potential}
T.~Kuroiwa, A.~Sarcon, T.~Ibara, E.~Yamada, A.~Yamamoto, K.~Tsukamoto, and K.~Fujita, ``The potential of chatgpt as a self-diagnostic tool in common orthopedic diseases: exploratory study,'' \emph{Journal of Medical Internet Research}, vol.~25, p. e47621, 2023.

\bibitem{liu2024timemattersexaminetemporal}
\BIBentryALTinterwordspacing
W.~Liu, Z.~He, and X.~Huang, ``Time matters: Examine temporal effects on biomedical language models,'' 2024. [Online]. Available: \url{https://arxiv.org/abs/2407.17638}
\BIBentrySTDinterwordspacing

\bibitem{lyu2022multimodal}
W.~Lyu, X.~Dong, R.~Wong, S.~Zheng, K.~Abell-Hart, F.~Wang, and C.~Chen, ``A multimodal transformer: Fusing clinical notes with structured ehr data for interpretable in-hospital mortality prediction,'' in \emph{AMIA Annual Symposium Proceedings}, vol. 2022.\hskip 1em plus 0.5em minus 0.4em\relax American Medical Informatics Association, 2022, p. 719.

\bibitem{li2024dalk}
D.~Li, S.~Yang, Z.~Tan, J.~Y. Baik, S.~Yun, J.~Lee, A.~Chacko, B.~Hou, D.~Duong-Tran, Y.~Ding \emph{et~al.}, ``Dalk: Dynamic co-augmentation of llms and kg to answer alzheimer's disease questions with scientific literature,'' \emph{arXiv preprint arXiv:2405.04819}, 2024.

\bibitem{hsu2024thought}
C.-Y. Hsu, K.~Cox, J.~Xu, Z.~Tan, T.~Zhai, M.~Hu, D.~Pratt, T.~Chen, Z.~Hu, and Y.~Ding, ``Thought graph: Generating thought process for biological reasoning,'' in \emph{Companion Proceedings of the ACM on Web Conference 2024}, 2024, pp. 537--540.

\bibitem{shi2022improving}
Y.~Shi, T.~ValizadehAslani, J.~Wang, P.~Ren, Y.~Zhang, M.~Hu, L.~Zhao, and H.~Liang, ``Improving imbalanced learning by pre-finetuning with data augmentation,'' in \emph{Proceedings of the Fourth International Workshop on Learning with Imbalanced Domains: Theory and Applications}, ser. Proceedings of Machine Learning Research, N.~Moniz, P.~Branco, L.~Torgo, N.~Japkowicz, M.~Wozniak, and S.~Wang, Eds., vol. 183.\hskip 1em plus 0.5em minus 0.4em\relax PMLR, 23 Sep 2022, pp. 68--82.

\bibitem{goldberger2000physiobank}
A.~L. Goldberger, L.~A. Amaral, L.~Glass, J.~M. Hausdorff, P.~C. Ivanov, R.~G. Mark, J.~E. Mietus, G.~B. Moody, C.-K. Peng, and H.~E. Stanley, ``Physiobank, physiotoolkit, and physionet: Components of a new research resource for complex physiologic signals,'' \emph{Circulation}, vol. 101, no.~23, pp. e215--e220, 2000.

\bibitem{mercuro2010gender}
G.~Mercuro, M.~Deidda, A.~Piras, C.~C. Dessalvi, S.~Maffei, and G.~M. Rosano, ``Gender determinants of cardiovascular risk factors and diseases,'' \emph{Journal of Cardiovascular Medicine}, vol.~11, no.~3, pp. 207--220, 2010.

\bibitem{quintero2012autoimmune}
O.~L. Quintero, M.~J. Amador-Patarroyo, G.~Montoya-Ortiz, A.~Rojas-Villarraga, and J.-M. Anaya, ``Autoimmune disease and gender: plausible mechanisms for the female predominance of autoimmunity,'' \emph{Journal of autoimmunity}, vol.~38, no. 2-3, pp. J109--J119, 2012.

\bibitem{li2023comparative}
Y.~Li, R.~M. Wehbe, F.~S. Ahmad, H.~Wang, and Y.~Luo, ``A comparative study of pretrained language models for long clinical text,'' \emph{Journal of the American Medical Informatics Association}, vol.~30, no.~2, pp. 340--347, 2023.

\bibitem{dixon2018measuring}
L.~Dixon, J.~Li, J.~Sorensen, N.~Thain, and L.~Vasserman, ``Measuring and mitigating unintended bias in text classification,'' in \emph{Proceedings of the AAAI/ACM Conference on AI, Ethics, and Society}, 2018, pp. 67--73.

\bibitem{snovaisg2023icd-mapping}
\BIBentryALTinterwordspacing
S.~Gon\c{c}alves and G.~Ehrensperger, ``Icd-mappings,'' 2023. [Online]. Available: \url{https://github.com/snovaisg/ICD-Mappings}
\BIBentrySTDinterwordspacing

\bibitem{wolf2020transformers}
\BIBentryALTinterwordspacing
T.~Wolf, L.~Debut, V.~Sanh, J.~Chaumond, C.~Delangue, A.~Moi, P.~Cistac, T.~Rault, R.~Louf, M.~Funtowicz, J.~Davison, S.~Shleifer, P.~von Platen, C.~Ma, Y.~Jernite, J.~Plu, C.~Xu, T.~Le~Scao, S.~Gugger, M.~Drame, Q.~Lhoest, and A.~Rush, ``Transformers: State-of-the-art natural language processing,'' in \emph{Proceedings of the 2020 Conference on Empirical Methods in Natural Language Processing: System Demonstrations}, Q.~Liu and D.~Schlangen, Eds.\hskip 1em plus 0.5em minus 0.4em\relax Online: Association for Computational Linguistics, Oct. 2020, pp. 38--45. [Online]. Available: \url{https://aclanthology.org/2020.emnlp-demos.6}
\BIBentrySTDinterwordspacing

\end{thebibliography}

\appendices

\section*{Appendix}
\label{2Appendix}

\subsection{Data Preprocessing}
\label{sub:appdix:data}

Our study examines imbalance effects on performance and fairness under phenotype inference task, predicting the top 100 most frequent International Classification of Diseases (ICD-10) codes from clinical notes. 
To accomplish this, we integrate data from the MIMIC-IV Notes and the MIMIC-IV database, combining discharge summaries, patient information (age, gender, ethnicity, insurance) and the diagnoses (ICD-10 codes). 
Our data preprocessing pipeline is:

\paragraph{Text Processing}
We extract discharge summaries from the MIMIC-IV-Notes~\cite{johnson2023mimic-note}.
Each summary undergoes data cleaning and tokenization. 
To ensure sufficient content for analysis, we exclude documents with fewer than 30 tokens.

\paragraph{Code Extraction and Mapping}
We collect diagnosis codes from the MIMIC-IV database, gathering both ICD-9 and ICD-10 codes for each hospital admission.
To ensure consistency across our label set, we convert ICD-9 codes to ICD-10. 
We keep the original ICD-10 for each note if available and convert the annotations into ICD-10 for notes that only have ICD-9 labels by the ICD-Mappings~\cite{snovaisg2023icd-mapping} toolkit.
We choose the top 100 frequent ICD-10 codes as the label set. 

\paragraph{Data Integration}
We use the unique identifier per subject to source patient information from the MIMIC-IV with the processed notes from MIMIC-IV-Note.
This includes gender, age, insurance type, and ethnicity/race, allowing to further analyze the data imbalances from these diverse perspectives.

\subsection{Data Imbalance}
\label{sub:appdix:imbalance}

\begin{table}[ht]
\centering
\caption{median age and age distribution of patients, by gender,   ethnicity or race, insurance.}
\resizebox{0.48\textwidth}{!}{
    \begin{tabular}{c|cccccc}
    & \textbf{Median}  & \textbf{18-29} & \textbf{30-49} & \textbf{50-69} & \textbf{70-89} & \textbf{90+} \\
    \hline \hline
    \textit{Gender} & & & & & & \\
    \hspace{0.2cm}Male & 61  & 7.61  & 20.40 & 36.09 & 31.05 & 4.85 \\
    Female & 61  & 5.46  & 18.25 & 43.74 & 29.73 & 2.81 \\
    \hline
    \textit{Ethnicity or Race}        &    &          &       &       &       &      \\
    \hspace{0.2cm}Hispanic/Latino        & 52  & 11.22 & 32.35 & 37.48 & 18.09 & 0.86 \\
    Asian & 60 & 9.52  & 20.98 & 36.85 & 30.13 & 2.51 \\
    Black & 57  & 8.41  & 25.23 & 40.91 & 23.43 & 2.03 \\
    White & 63 & 5.39  & 16.78 & 40.12 & 33.12 & 4.59 \\
    Other & 59 & 8.95 & 21.38	& 38.13	& 28.11	& 3.42\\
    \hline
    \textit{Insurance }                &       &       &       &       &       &      \\
    \hspace{0.2cm}Medicaid                  & 48  & 13.49 & 37.02 & 42.51 & 6.59  & 0.39 \\
    Medicare & 72  & 1.03  & 8.04  & 32.06 & 51.44 & 7.44 \\
    Other & 55  & 9.92  & 25.68 & 45.70 & 17.19 & 1.51\\
    \end{tabular}
}
\label{tab: age distribution}
\end{table}

\paragraph{Ethnicity Imbalance}

The data exhibits significant ethnicity imbalance. 
The White population predominates, comprising 68.90\% of the data, as shown in the first column of Table~\ref{tab: gender and insurance distribution by race}. 
Black/African American individuals form the second-largest group, accounting for 14.77\%, while Hispanic/Latino and Asian patients count only 5.28\% and 3.21\%, respectively.

\paragraph{Gender Imbalance}
The overall gender distribution is slightly imbalanced, with 52.08\% females and 48.92\% males. 
But when considering gender distribution inside specific ethnic groups, the gender distribution can be very imbalanced. 
For example, the female ratio is  60.37\% in the Black/African American group, comparing to 49.37\% (very balanced) in the White group as shown in Table~\ref{tab: gender and insurance distribution by race}.

\begin{table}[ht]
\centering
\caption{Gender and insurance distribution by Ethnicity and Race}
\resizebox{.485\textwidth}{!}{
\begin{tabular}{l|c|cc|ccc}
& & \multicolumn{2}{c|}{Gender (\%)} & \multicolumn{3}{c}{Insurance (\%)} \\
Ethnicity\&Race & Total & F & M & Medicaid & Medicare  & Other \\
\hline \hline
Hispanic/Latino & 5.28  & 52.37 & 47.63 & 22.48 & 27.03 & 50.49 \\
Asian & 3.21  & 50.57 & 49.43 & 16.42 & 22.28 & 61.30 \\
Black & 14.77 & 60.37 & 39.63 & 13.18 & 36.16 & 50.66 \\
White & 68.90 & 49.38 & 50.62 & 5.34 & 45.07 & 49.59 \\
Other & 7.84 & 47.83 & 52.17 & 9.69 & 32.66	& 57.64\\
\hline
All Data & 100 & 51.08 & 48.92 & 8.10 & 41.10 & 50.80 \\
\end{tabular}
}
\label{tab: gender and insurance distribution by race}
\end{table}
\paragraph{Age Group Imbalance}
The dataset exhibits significant imbalance across age groups. 
Table~\ref{tab: age distribution} presents the median age and age distribution of patients, categorized by gender, ethnicity or race, and Social Determinants of Health (SDOH), such as insurance type.
The dataset contains no samples under 17 years of age. 
The majority of the data is concentrated in the 50-69 and 70-89 age groups across most demographic categories. 
The median age for both males and females is 61 years, indicating a skew towards older populations.

\paragraph{Social Determinants of Health Imbalance}
Insurance is one of the social determinants of health.
The patient insurance distribution of the dataset also demonstrates imbalance.
As shown in Table~\ref{tab: gender and insurance distribution by race}, the Other insurance category is most prevalent overall at 50.80\%, followed by Medicare at 41.10\%, and Medicaid at 8.10\%. 
Additionally, this distribution varies considerably across ethnic and racial groups. 
The White population closely mirrors the overall distribution, while other groups show marked differences. 
Hispanic or Latino patients have the highest proportion of Medicaid coverage (22.48\%), which is more than four times the overall average. 
Asian patients have the highest proportion in the Other insurance category (61.30\%). Black or African American patients show a more balanced distribution between Other insurance and Medicare, but still have higher than average Medicaid coverage. 

\paragraph{Label Imbalance}
For our phenotype inference task, we employ ICD-10 codes as labels, selecting the top 100 most frequent codes to form our label set. 
The label distribution is notably imbalanced, with the frequency of individual ICD-10 codes in the set ranging from 2.19\% to 38.87\%.

\subsection{Implementation Details}
\label{sub:appdix:implementation} 
We implemented our experiments by Huggingface Transformers~\cite{wolf2020transformers} and PyTorch. 
Our experiments trained the ClinicalBERT and GatorTron models with a maximum sequence length of 512 tokens for the discharge summaries.
For the Clinical Longformer model, which is designed to handle longer text sequences, we used the maximum supported sequence length of 4,096 tokens.
All models were trained over 10 epochs with the same learning rate of $2 \times 10^{-5}$ and maximized batch sizes to fit NVIDIA 4090 GPU memory.

\end{document}